\documentclass{article} 
\usepackage{nips11submit_e,times}
\usepackage{cite}
\usepackage[pdftex]{graphicx}
\usepackage{rotating}
\usepackage{amsmath}
\usepackage{amssymb}
\usepackage{array}
\usepackage{nth}
\makeatletter

\newcommand{\Rmnum}[1]{\expandafter\@slowromancap\romannumeral #1@}
\makeatother

\title{Additive Gaussian Processes}

\author{
David Duvenaud \\
Department of Engineering\\
Cambridge University\\
\texttt{dkd23@cam.ac.uk} \\
\And
Hannes Nickisch \\
MPI for Intelligent Systems \\
T\"{u}bingen, Germany \\
\texttt{hn@tue.mpg.de} \\
\And
Carl Edward Rasmussen \\
Department of Engineering \\
Cambridge University \\
\texttt{cer54@cam.ac.uk} \\
}

\nipsfinalcopy 
\begin{document}
\maketitle

\begin{abstract}
We introduce a Gaussian process model of functions which are $\textit additive$.  An additive function is one which decomposes into a sum of low-dimensional functions, each depending on only a subset of the input variables. Additive GPs generalize both Generalized Additive Models, and the standard GP models which use squared-exponential kernels.  Hyperparameter learning in this model can be seen as Bayesian Hierarchical Kernel Learning (HKL).  We introduce an expressive but tractable parameterization of the kernel function, which allows efficient evaluation of all input interaction terms, whose number is exponential in the input dimension.  The additional structure discoverable by this model results in increased interpretability, as well as state-of-the-art predictive power in regression tasks.
\end{abstract}

\section{Introduction}
Most statistical regression models in use today are of the form: $g(y) = f(x_1) + f(x_2) + \dots + f(x_D)$.  Popular examples include logistic regression, linear regression, and Generalized Linear Models\cite{nelder1972generalized}.  This family of functions, known as Generalized Additive Models (GAM)\cite{hastie1990generalized}, are typically easy to fit and interpret.   Some extensions of this family, such as smoothing-splines ANOVA \cite{wahba1990spline}, add terms depending on more than one variable.  However, such models generally become intractable and difficult to fit as the number of terms increases.

At the other end of the spectrum are kernel-based models, which typically allow the response to depend on all input variables simultaneously.  These have the form: $y = f(x_1, x_2, \dots, x_D)$.  A popular example would be a Gaussian process model using a squared-exponential (or Gaussian) kernel.  We denote this model as SE-GP.  This model is much more flexible than the GAM, but its flexibility makes it difficult to generalize to new combinations of input variables. 

In this paper, we introduce a Gaussian process model that generalizes both GAMs and the SE-GP.  This is achieved through a kernel which allow additive interactions of all orders, ranging from first order interactions (as in a GAM) all the way to $D$th-order interactions (as in a SE-GP).  Although this kernel amounts to a sum over an exponential number of terms, we show how to compute this kernel efficiently, and introduce a parameterization which limits the number of hyperparameters to $O(D)$.  A Gaussian process with this kernel function (an additive GP) constitutes a powerful model that allows one to automatically determine which orders of interaction are important.  We show that this model can significantly improve modeling efficacy, and has major advantages for model interpretability.  This model is also extremely simple to implement, and we provide example code.

We note that a similar breakthrough has recently been made, called Hierarchical Kernel Learning (HKL)\cite{DBLP:journals/corr/abs-0909-0844}.  HKL explores a similar class of models, and sidesteps the possibly exponential number of interaction terms by cleverly selecting only a tractable subset.  However, this method suffers considerably from the fact that cross-validation must be used to set hyperparameters.  In addition, the machinery necessary to train these models is immense.  Finally, on real datasets, HKL is outperformed by the standard SE-GP \cite{DBLP:journals/corr/abs-0909-0844}.

\section{Gaussian Process Models}

Gaussian processes are a flexible and tractable prior over functions, useful for solving regression and classification tasks\cite{rasmussen38gaussian}.  The kind of structure which can be captured by a GP model is mainly determined by its \emph{kernel}: the covariance function.  One of the main difficulties in specifying a Gaussian process model is in choosing a kernel which can represent the structure present in the data.  For small to medium-sized datasets, the kernel has a large impact on modeling efficacy.

\begin{figure}
\centering
\begin{tabular}{ccccc|c}
\hspace{-0.2cm}\includegraphics[width=0.2\textwidth]{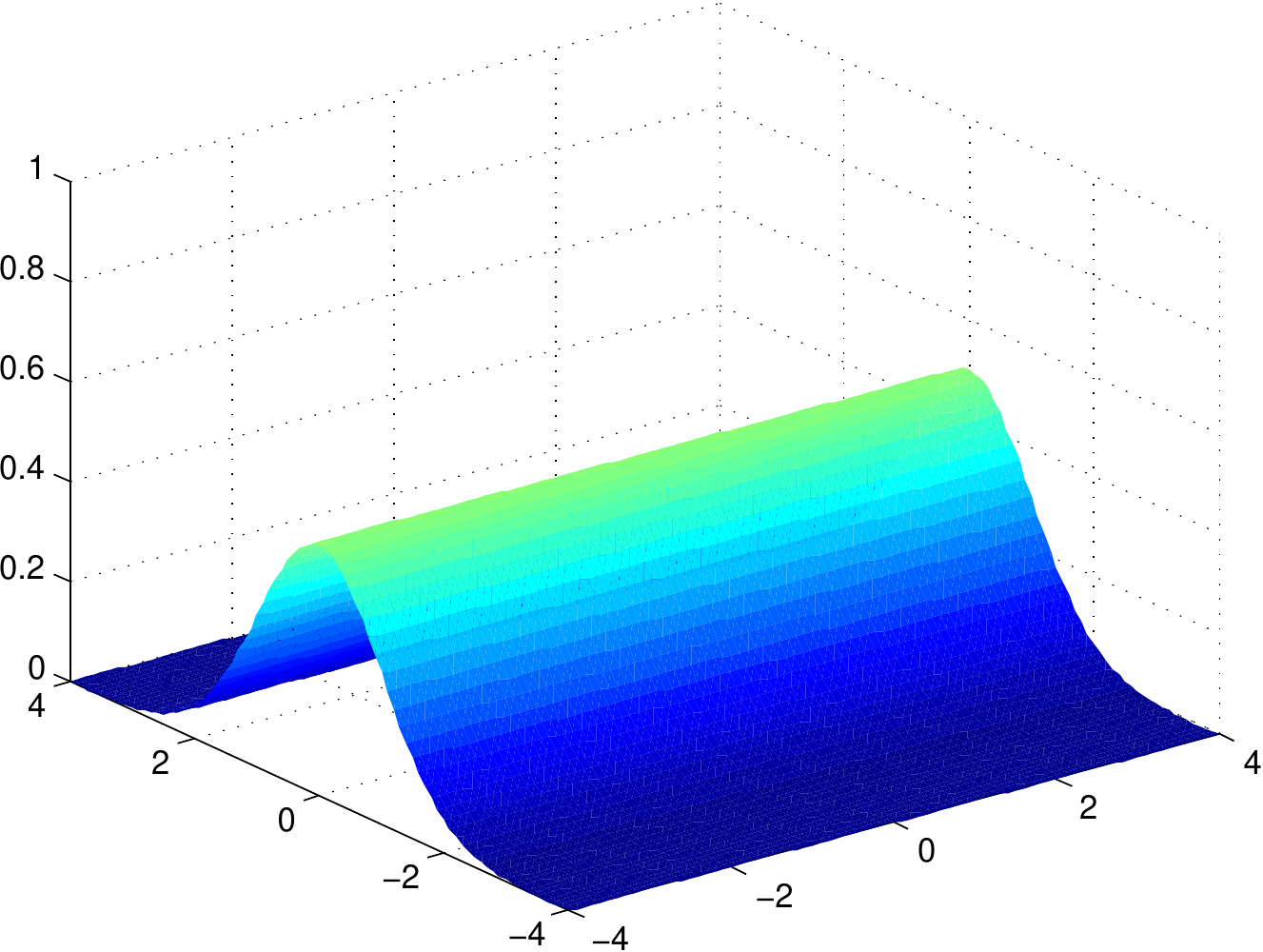} & \hspace{-0.4cm} + \hspace{-0.4cm} & 
\includegraphics[width=0.2\textwidth]{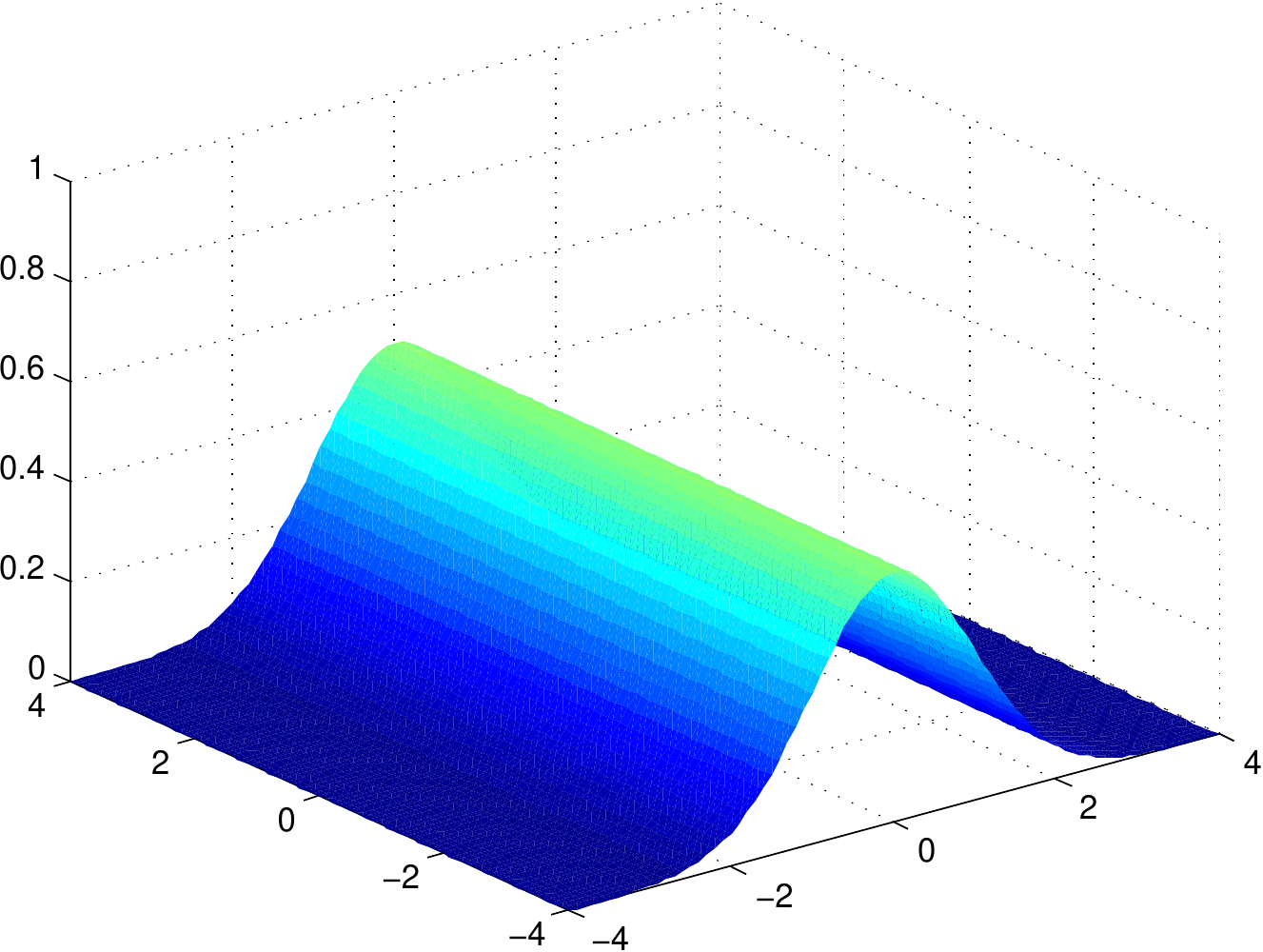} & \hspace{-0.4cm} = \hspace{-0.4cm} & 
\includegraphics[width=0.2\textwidth]{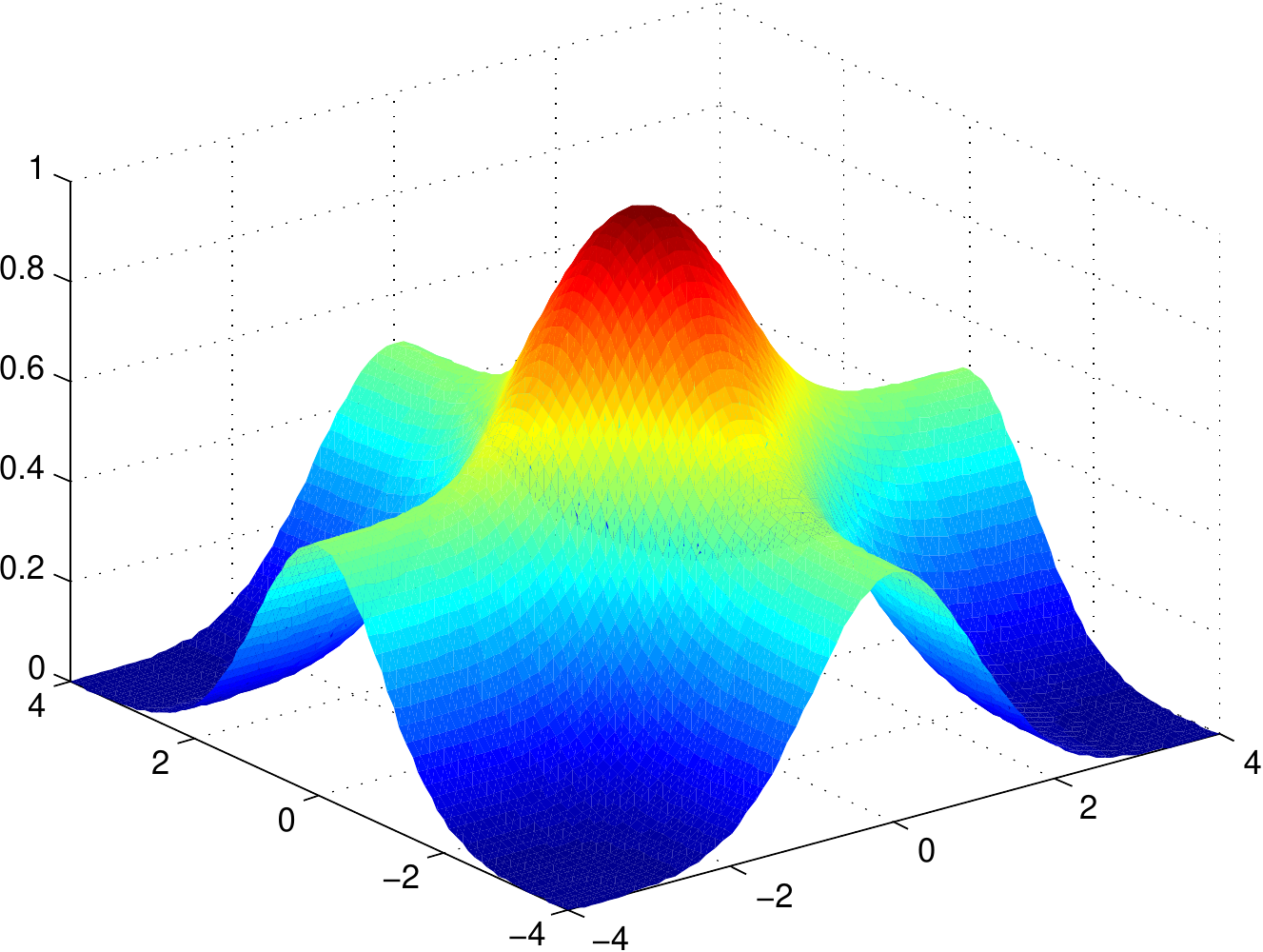} &
\includegraphics[width=0.2\textwidth]{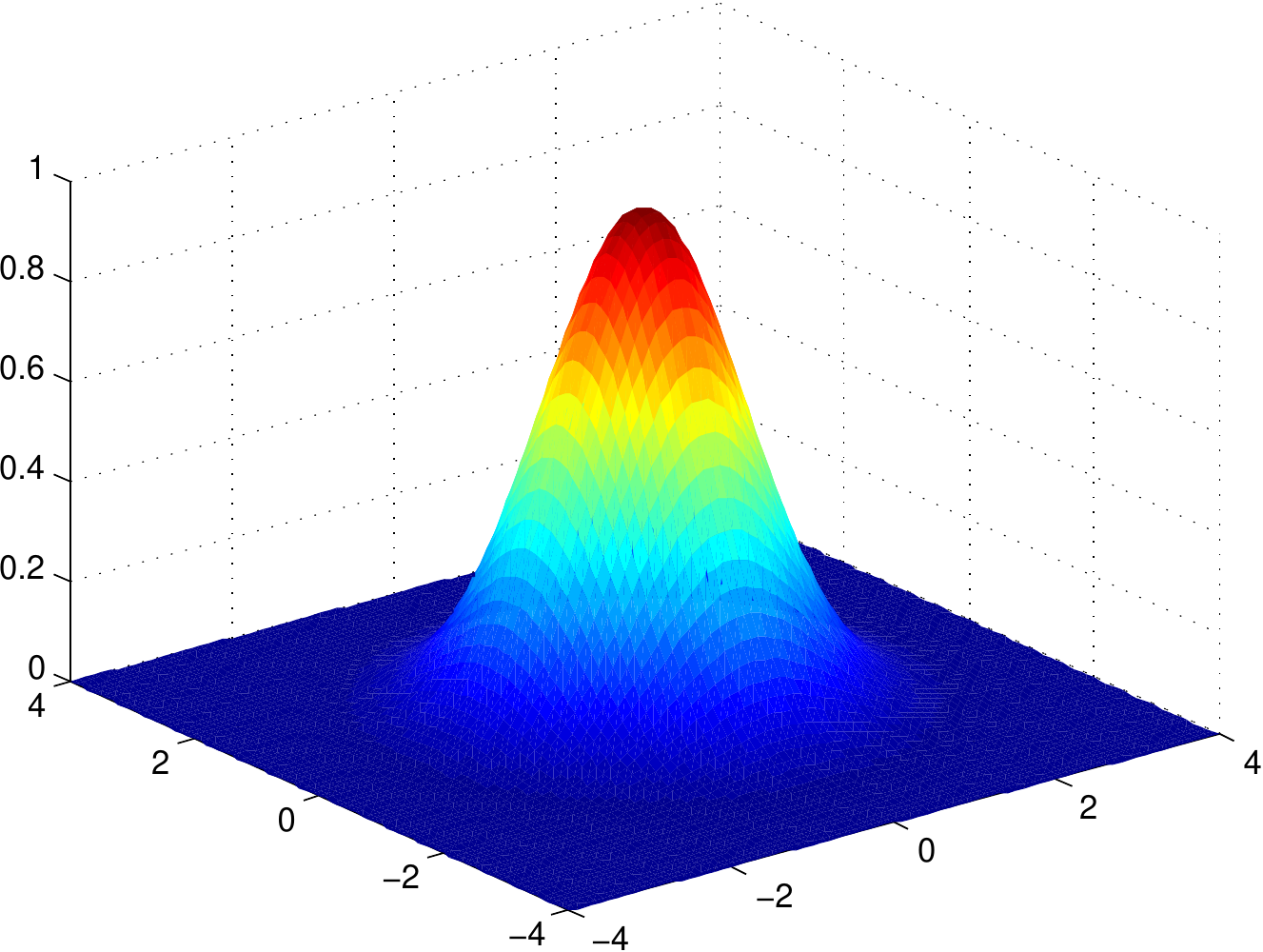} \\
$k_1(x_1, x_1')$ & & $k_2(x_2, x_2')$ & & $k_1(x_1,x_1') + k_2(x_2,x_2')$ &$k_1(x_1,x_1')k_2(x_2,x_2')$ \\
1D kernel & & 1D kernel & & 1st order kernel & 2nd order kernel \\ 
$\downarrow$ & & $\downarrow$ & & $\downarrow$ & $\downarrow$  \\
\hspace{-0.2cm}\includegraphics[width=0.2\textwidth]{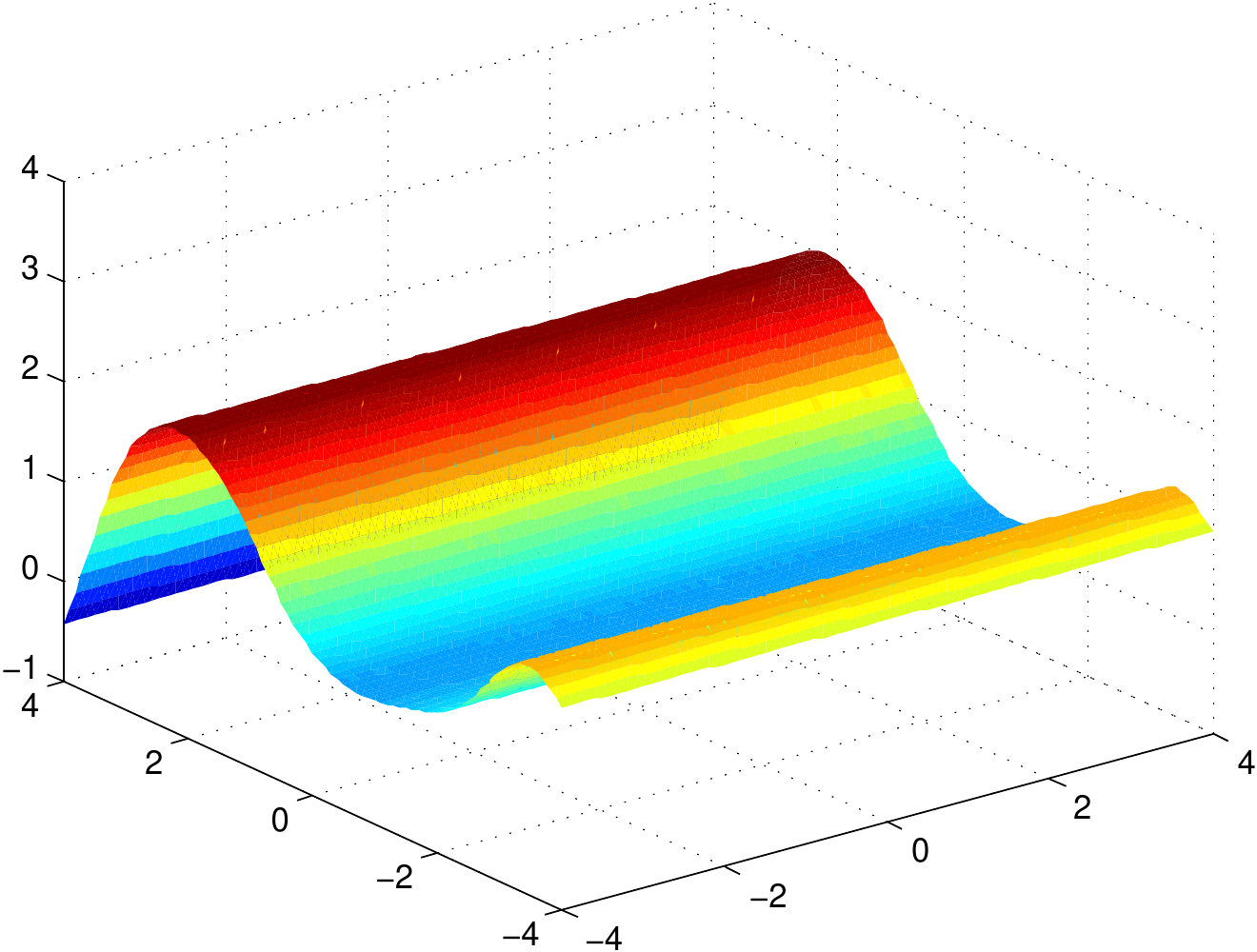}& \hspace{-0.4cm} + \hspace{-0.4cm}& 
\includegraphics[width=0.2\textwidth]{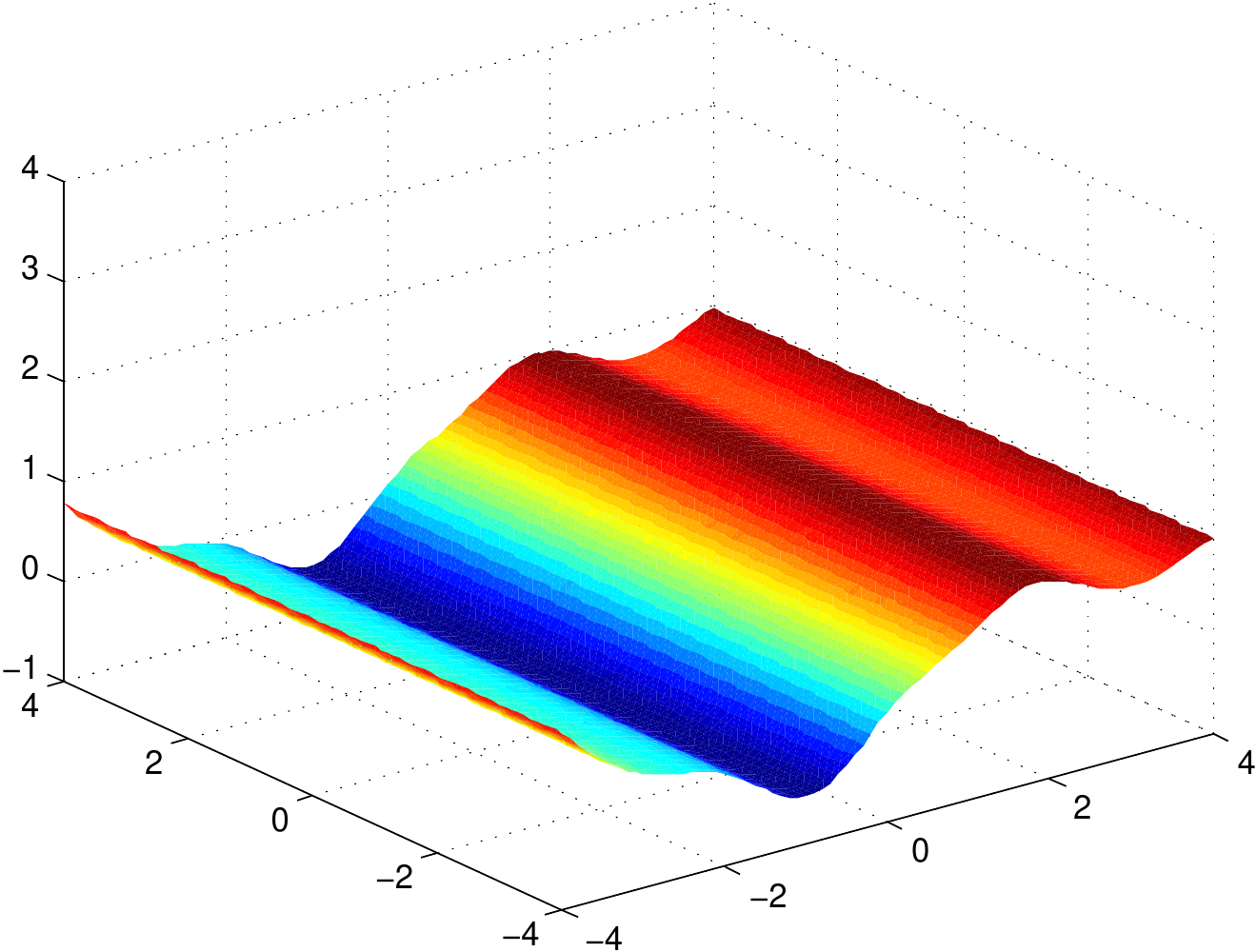}& \hspace{-0.4cm} = \hspace{-0.4cm}&
\includegraphics[width=0.2\textwidth]{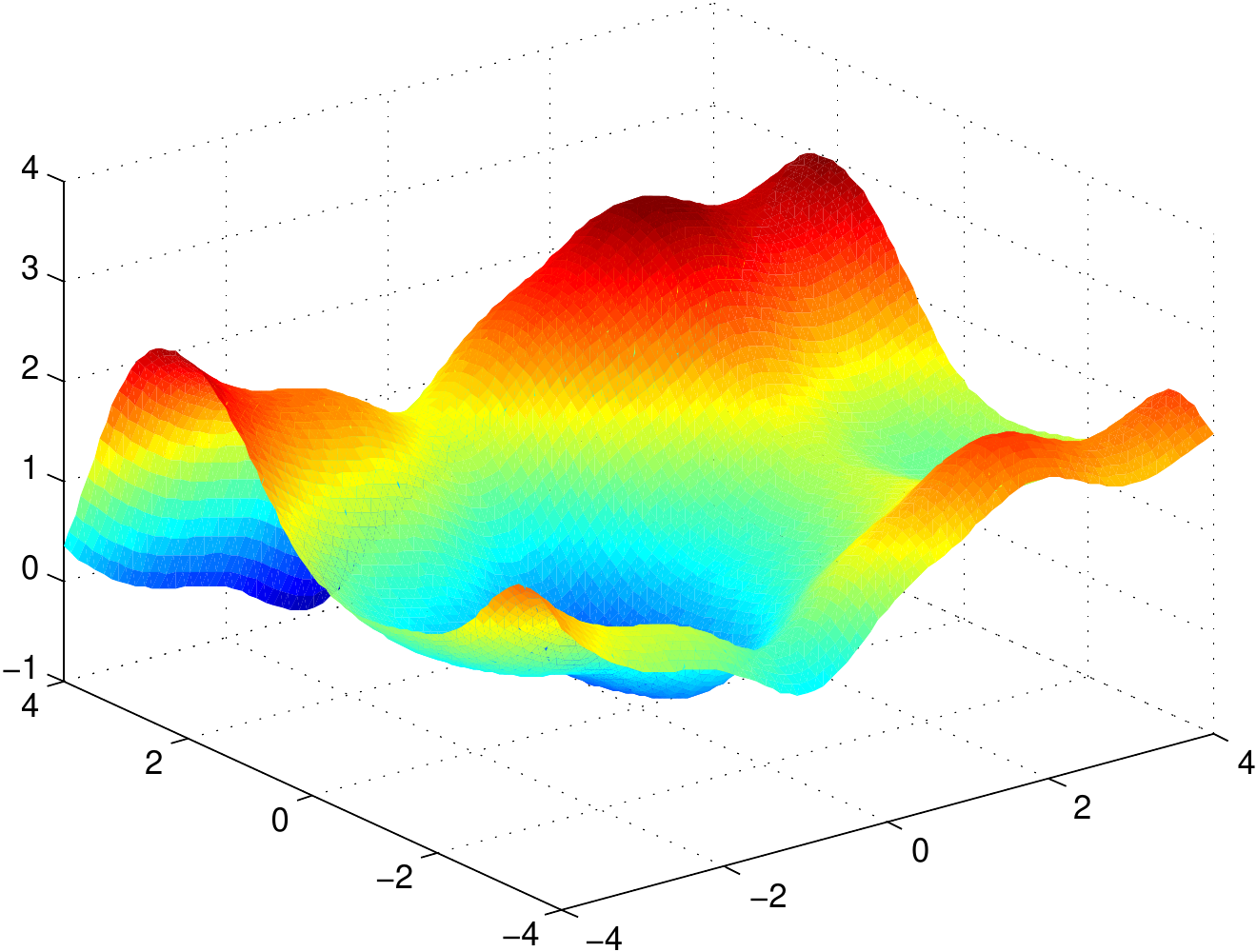} &
\includegraphics[width=0.2\textwidth]{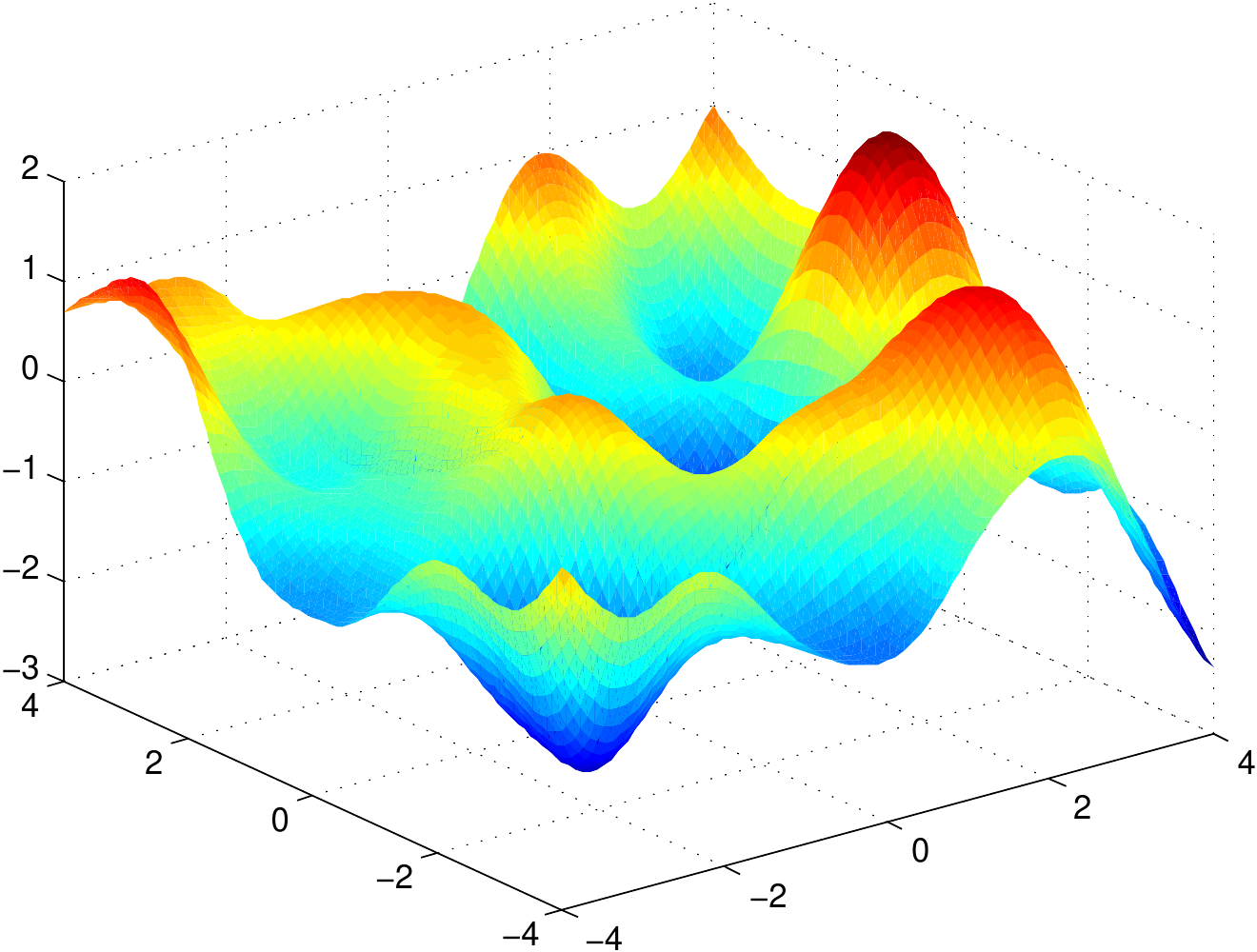} \\
$f_1(x_1)$ & & $f_2(x_2)$ & & $f_1(x_1) + f_2(x_2)$ & $f(x_1, x_2)$ \\
draw from & & draw from & & draw from & draw from\\
1D GP prior & & 1D GP prior & & 1st order GP prior & 2nd order GP prior\\
\end{tabular}
\caption{A first-order additive kernel, and a product kernel.  Left: a draw from a first-order additive kernel corresponds to a sum of draws from one-dimensional kernels.  Right: functions drawn from a product kernel prior have weaker long-range dependencies, and less long-range structure.
}
\label{fig:kernels}
\end{figure}

Figure \ref{fig:kernels} compares, for two-dimensional functions, a first-order additive kernel with a second-order kernel. We can see that a GP with a first-order additive kernel is an example of a GAM:  Each function drawn from this model is a sum of orthogonal one-dimensional functions.  Compared to functions drawn from the higher-order GP, draws from the first-order GP have more long-range structure.



We can expect many natural functions to depend only on sums of low-order interactions.  For example, the price of a house or car will presumably be well approximated by a sum of prices of individual features, such as a sun-roof.  
Other parts of the price may depend jointly on a small set of features, such as the size and building materials of a house.
Capturing these regularities will mean that a model can confidently extrapolate to unseen combinations of features.




\section{Additive Kernels}

We now give a precise definition of additive kernels.  We first assign each dimension $i \in \{1 \dots D\}$ a one-dimensional \emph{base kernel} $k_i(x_i, x'_i)$.  We then define the first order, second order and $n$th order additive kernel as:
\begin{eqnarray}
k_{add_1}({\bf x, x'}) & = & \sigma_1^2 \sum_{i=1}^D k_i(x_i, x_i') \\
k_{add_2}({\bf x, x'}) & = & \sigma_2^2 \sum_{i=1}^D \sum_{j = i + 1}^D k_i(x_i, x_i') k_j(x_j, x_j') \\
k_{add_n}({\bf x, x'}) & = & \sigma_n^2 \sum_{1 \leq i_1 < i_2 < ... < i_n \leq D} \left[ \prod_{d=1}^n k_{i_d}(x_{i_d}, x_{i_d}') \right]
\end{eqnarray}
where $D$ is the dimension of our input space, and $\sigma_n^2$ is the variance assigned to all $n$th order interactions. The $n$th covariance function is a sum of ${D \choose n}$ terms.  In particular, the $D$th order additive covariance function has ${D \choose D} = 1$ term, a product of each dimension's covariance function:
\begin{equation}
k_{add_D}({\bf x, x'}) = \sigma_D^2 \prod_{d=1}^D k_{d}(x_{d}, x_{d}')
\end{equation}
In the case where each base kernel is a one-dimensional squared-exponential kernel, the $D$th-order term corresponds to the multivariate squared-exponential kernel:
\begin{equation}
k_{add_D}({\bf x, x'}) = \sigma_D^2 \prod_{d=1}^D k_{d}(x_{d}, x_{d}') = \sigma_D^2 \prod_{d=1}^D \exp \Big( -\frac{ ( x_{d} - x_{d}')^2}{2l^2_d} \Big) = \sigma_D^2  \exp \Big( -\sum_{d=1}^D \frac{ ( x_{d} - x_{d}')^2}{2l^2_d} \Big)
\end{equation}
also commonly known as the Gaussian kernel.  The full additive kernel is a sum of the additive kernels of all orders.
\subsection{Parameterization}

The only design choice necessary in specifying an additive kernel is the selection of a one-dimensional base kernel for each input dimension.  Any parameters (such as length-scales) of the base kernels can be learned as usual by maximizing the marginal likelihood of the training data.  

In addition to the hyperparameters of each dimension-wise kernel, additive kernels are equipped with a set of $D$ hyperparameters $\sigma_1^2 \dots \sigma_D^2$ controlling how much variance we assign to each order of interaction.  These ``order variance'' hyperparameters have a useful interpretation:  The $d$th order variance hyperparameter controls how much of the target function's variance comes from interactions of the $d$th order.
%
%
%
Table \ref{tbl:all_orders} shows examples of normalized order variance hyperparameters learned on real datasets.
\begin{table}[h]
\caption{{\small
Relative variance contribution of each order in the additive model, on different datasets. Here, the maximum order of interaction is set to 10, or smaller if the input dimension less than 10.  Values are normalized to sum to 100.
}}
\label{tbl:all_orders}
\begin{center}
\begin{tabular}{r | r r r r r r r r r r}
Order of interaction & \nth{1} & \nth{2} & \nth{3} & \nth{4} & \nth{5} & \nth{6} & \nth{7} & \nth{8} & \nth{9} & \nth{10} \\ \hline
pima  & $0.1 $ & $0.1 $ & $0.1 $ & $0.3 $ & $1.5 $ & ${\bf 96.4}$ & $1.4 $ & $0.0 $ & & \\
liver  & $0.0 $ & $0.2 $ & ${\bf 99.7 } $ & $0.1 $ & $0.0 $ & $0.0 $ & & & & \\
heart  & ${\bf 77.6} $ & $0.0 $ & $0.0 $ & $0.0 $ & $0.1 $ & $0.1 $ & $0.1 $ & $0.1 $ & $0.1 $ & $22.0 $ \\
concrete  & ${\bf 70.6 } $ & $13.3 $ & $13.8 $ & $2.3 $ & $0.0 $ & $0.0 $ & $0.0 $ & $0.0 $ & & \\
pumadyn-8nh  & $0.0 $ & $0.1 $ & $0.1 $ & $0.1 $ & $0.1 $ & $0.1 $ & $0.1 $ & ${\bf 99.5 } $ & & \\
servo  & ${\bf 58.7 }$ & $27.4 $ & $0.0 $ & $13.9 $ & & & & & & \\
housing  & $0.1 $ & $0.6 $ & ${\bf 80.6 }$ & $1.4 $ & $1.8 $ & $0.8 $ & $0.7 $ & $0.8 $ & $0.6 $ & $12.7 $ \\
\end{tabular}
\end{center}
\end{table}

On different datasets, the dominant order of interaction estimated by the additive model varies widely.  An additive GP with all of its variance coming from the 1st order is equivalent to a GAM; an additive GP with all its variance coming from the $D$th order is equivalent to a SE-GP.
%
%


Because the hyperparameters can specify which degrees of interaction are important, the additive GP is an extremely general model.  
  If the function we are modeling is decomposable into a sum of low-dimensional functions, our model can discover this fact and exploit it (see Figure \ref{fig:synth2d}) .  If this is not the case, the hyperparameters can specify a suitably flexible model.


\subsection{Interpretability}

\begin{figure}[h]
\centering
\begin{tabular}{ccc}
\includegraphics[width=0.3\textwidth]{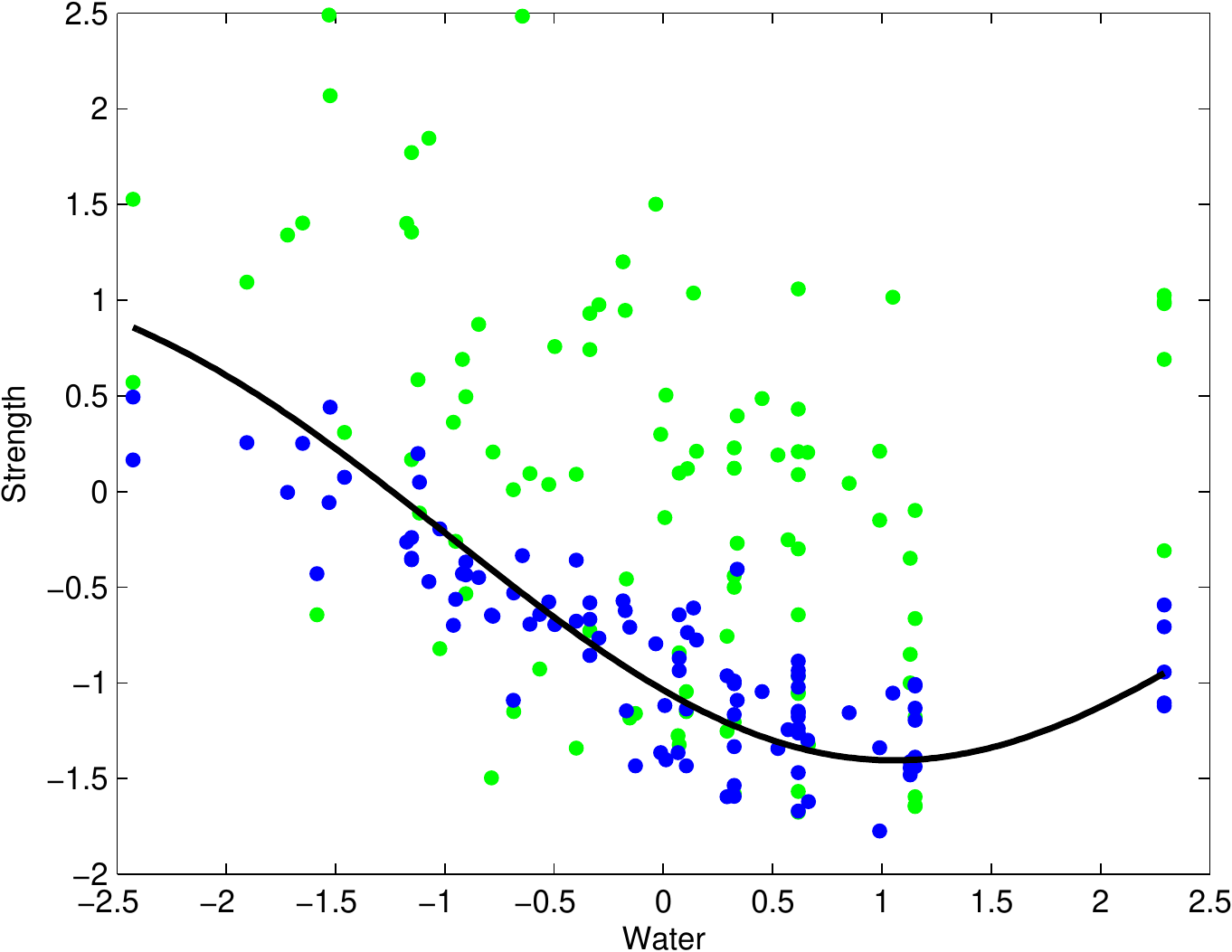} &
\includegraphics[width=0.3\textwidth]{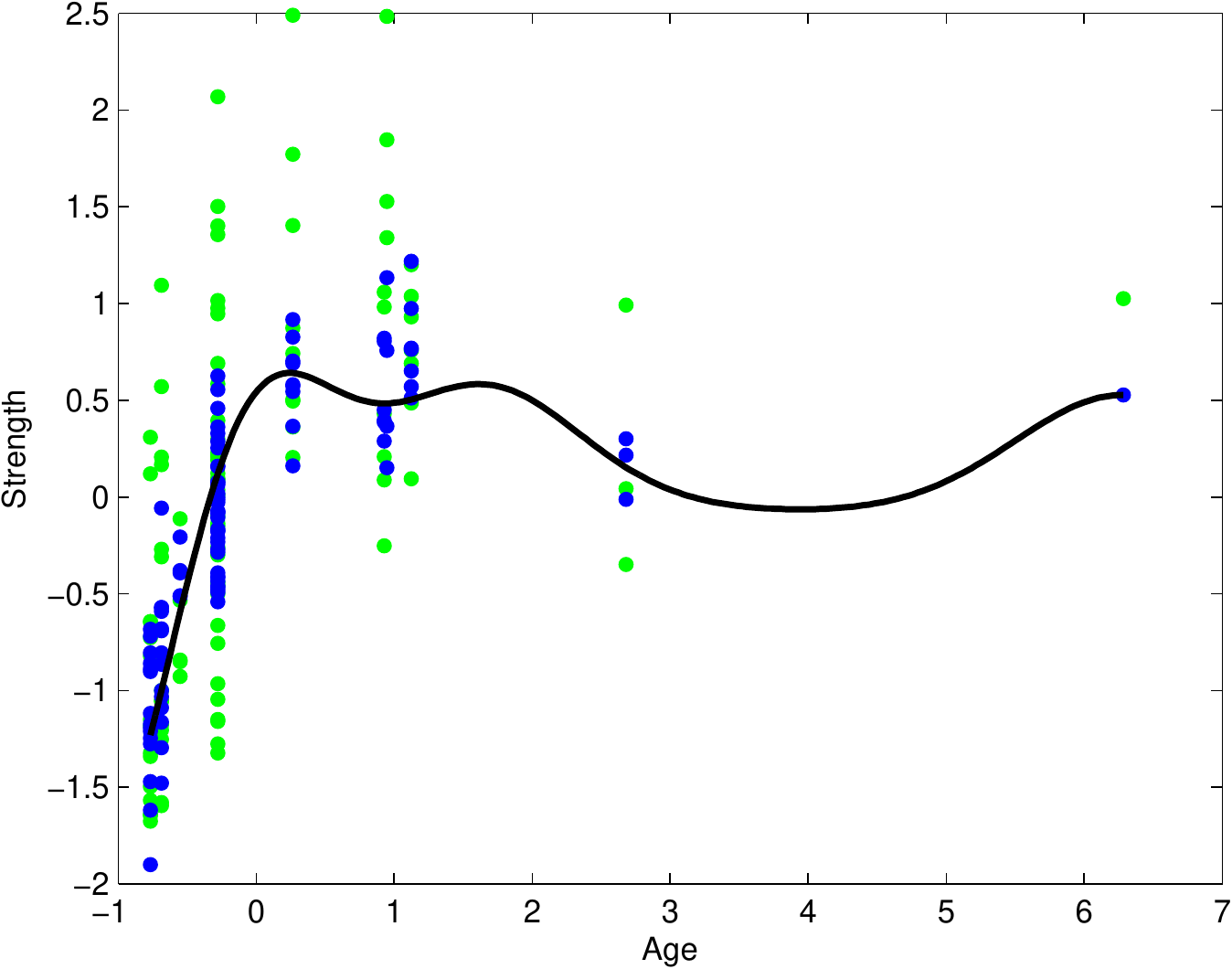}& 
\includegraphics[width=0.3\textwidth]{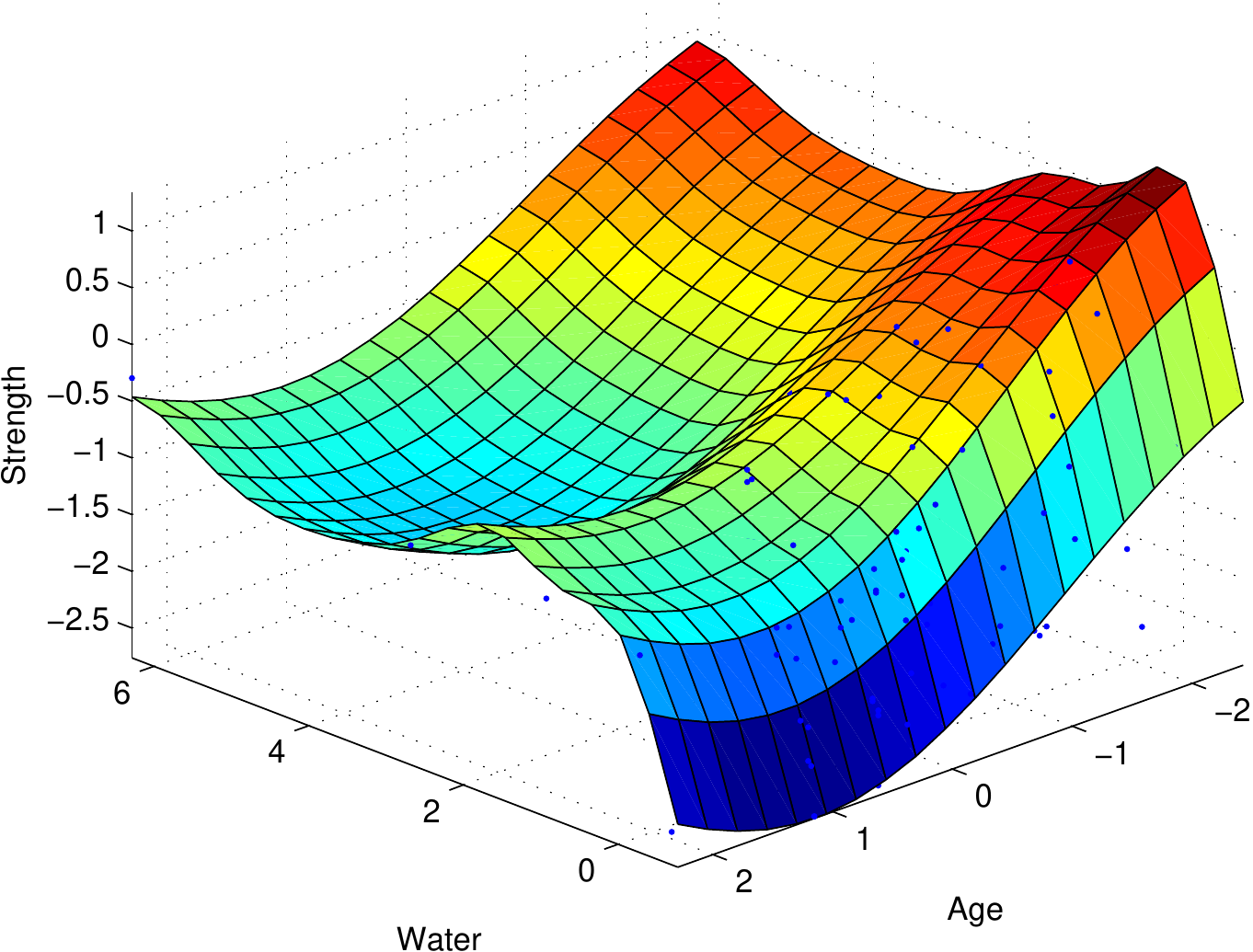}\\
\end{tabular}
\caption{Low-order functions on the concrete dataset.  Left, Centre:  By considering only first-order terms of the additive kernel, we recover a form of Generalized Additive Model, and can plot the corresponding 1-dimensional functions.  Green points indicate the original data, blue points are data after the mean contribution from the other dimensions' first-order terms has been subtracted.  The black line is the posterior mean of a GP with only one term in its kernel.  Right:  The posterior mean of a GP with only one second-order term in its kernel.}
\label{fig:interpretable functions}
\end{figure}

As noted by Plate\cite{plate1999accuracy}, one of the chief advantages of additive models such as GAM is their interpretability.
Plate also notes that by allowing high-order interactions as well as low-order interactions, one can trade off interpretability with predictive accuracy.  In the case where the hyperparameters indicate that most of the variance in a function can be explained by low-order interactions, it is useful and easy to plot the corresponding low-order functions, as in Figure \ref{fig:interpretable functions}. 

\subsection{Efficient Evaluation of Additive Kernels}
An additive kernel over $D$ inputs with interactions up to order $n$ has $O(2^n)$ terms.  Na\"{i}vely summing over these terms quickly becomes intractable.  In this section, we show how one can evaluate the sum over all terms in $O(D^2)$.

The $n$th order additive kernel corresponds to the $n$th \textit{elementary symmetric polynomial}\cite{macdonald1998symmetric} \cite{stanley2001enumerative}, which we denote $e_n$.  For example:  if $\bf x$ has 4 input dimensions ($D = 4$), and if we let $z_i = k_i(x_i,x_i')$, then
\begin{align*}
k_{add_1}({\bf x, x'}) & = e_1( z_1, z_2, z_3, z_4 ) = z_1 + z_2 + z_3 + z_4 \\
k_{add_2}({\bf x, x'}) & = e_2( z_1, z_2, z_3, z_4 ) = z_1 z_2 + z_1 z_3 + z_1z_4 + z_2 z_3 + z_2 z_4 + z_3 z_4 \\
k_{add_3}({\bf x, x'}) & = e_3( z_1, z_2, z_3, z_4 ) = z_1 z_2 z_3 + z_1 z_2 z_4 + z_1 z_3 z_4 + z_2 z_3 z_4 \\
k_{add_4}({\bf x, x'}) & = e_4( z_1, z_2, z_3, z_4 ) = z_1 z_2 z_3 z_4
\end{align*}
The Newton-Girard formulae give an efficient recursive form for computing these polynomials.  If we define $s_k$ to be the $k$th power sum:  $s_k(z_1,z_2,\dots,z_D) = \sum_{i=1}^Dz_i^k$, then
\begin{equation}
k_{add_n}({\bf x, x'}) = e_n(z_1,\dots,z_D) = \frac{1}{n} \sum_{k=1}^n (-1)^{(k-1)} e_{n-k}(z_1,\dots,z_D)s_k(z_1,\dots,z_D)
\end{equation}
Where $e_0 \triangleq 1$.  The Newton-Girard formulae have time complexity $O( D^2 )$, while computing a sum over an exponential number of terms.

Conveniently, we can use the same trick to efficiently compute all of the necessary derivatives of the additive kernel with respect to the base kernels.  We merely need to remove the kernel of interest from each term of the polynomials:
\begin{align}
\frac{\partial k_{add_n}}{\partial z_j} & = e_{n-1}(z_1,\dots,z_{j-1},z_{j+1}, \dots z_D)
\end{align}
This trick allows us to optimize the base kernel hyperparameters with respect to the marginal likelihood.

\subsection{Computation}
The computational cost of evaluating the Gram matrix of a product kernel (such as the SE kernel) is $O(N^2D)$, while the cost of evaluating the Gram matrix of the additive kernel is $O(N^2DR)$, where R is the maximum degree of interaction allowed (up to D).  In higher dimensions, this can be a significant cost, even relative to the fixed $O(N^3)$ cost of inverting the Gram matrix.
However, as our experiments show, typically only the first few orders of interaction are important for modeling a given function; hence if one is computationally limited, one can simply limit the maximum degree of interaction without losing much accuracy.

Additive Gaussian processes are particularly appealing in practice because their use requires only the specification of the base kernel.  All other aspects of GP inference remain the same.  All of the experiments in this paper were performed using the standard GPML toolbox\footnote{Available at \texttt{http://www.gaussianprocess.org/gpml/code/}}; code to perform all experiments is available at the author's website.\footnote{Example code available at: \texttt{http://mlg.eng.cam.ac.uk/duvenaud/}}

\section{Related Work}

Plate\cite{plate1999accuracy} constructs a form of additive GP, but using only the first-order and $D$th order terms.  This model is motivated by the desire to trade off the interpretability of first-order models, with the flexibility of full-order models.  Our experiments show that often, the intermediate degrees of interaction contribute most of the variance.

A related functional ANOVA GP model\cite{kaufman2010bayesian} decomposes the \emph{mean} function into a weighted sum of GPs. However, the effect of a particular degree of interaction cannot be quantified by that approach. Also, computationally, the Gibbs sampling approach used in \cite{kaufman2010bayesian} is disadvantageous.


Christoudias et al.\cite{christoudias2009bayesian} previously showed how mixtures of kernels can be learnt by gradient descent in the Gaussian process framework.  They call this \emph{Bayesian localized multiple kernel learning}.
However, their approach learns a mixture over a small, fixed set of kernels, while our method learns a mixture over all possible products of those kernels.

\subsection{Hierarchical Kernel Learning}

\begin{figure}
\centering
\begin{tabular}{c|c|c|c}
\hspace{-0.06in} \includegraphics[width=0.23\textwidth]{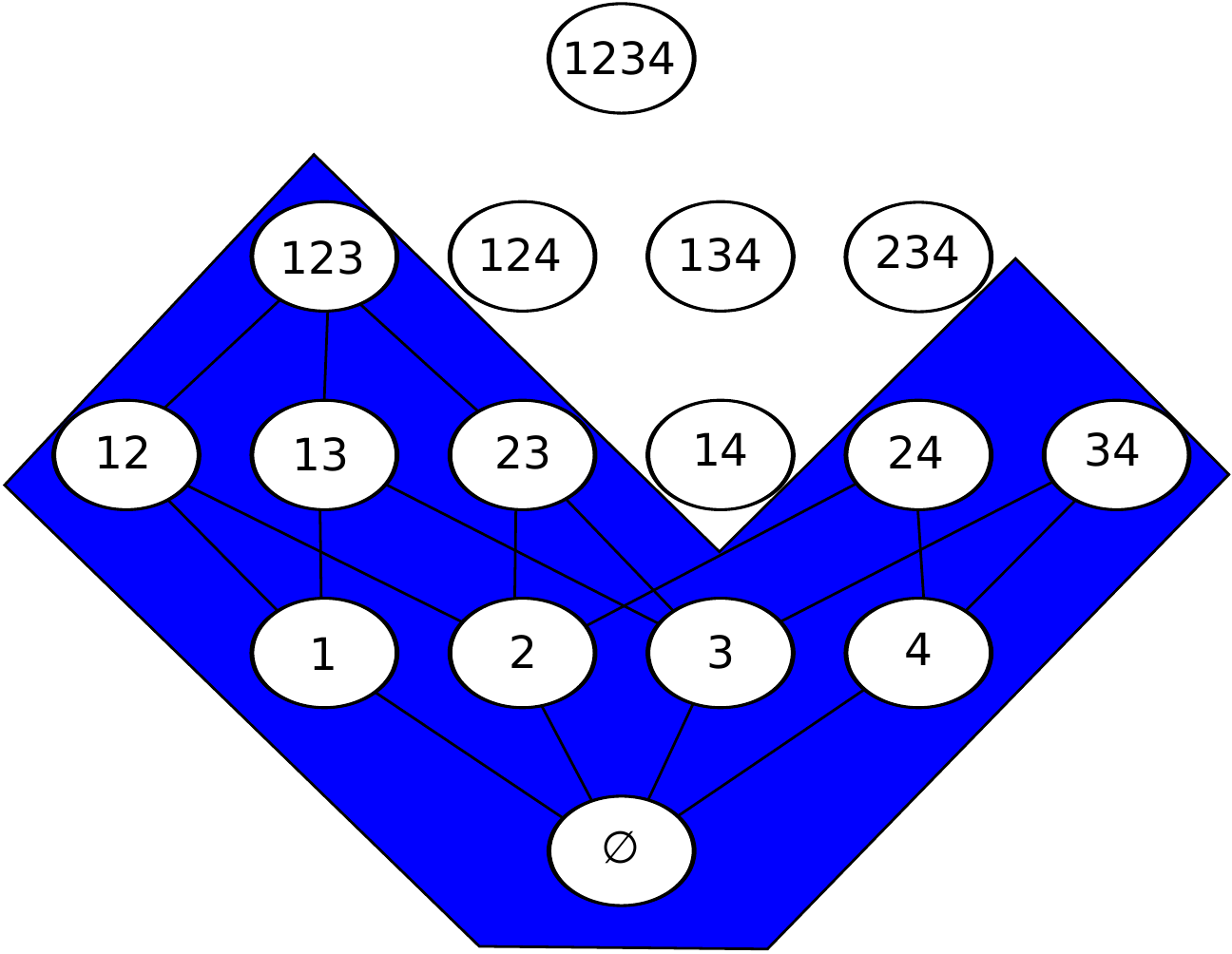} \hspace{-0.07in} &
\hspace{-0.06in} \includegraphics[width=0.23\textwidth]{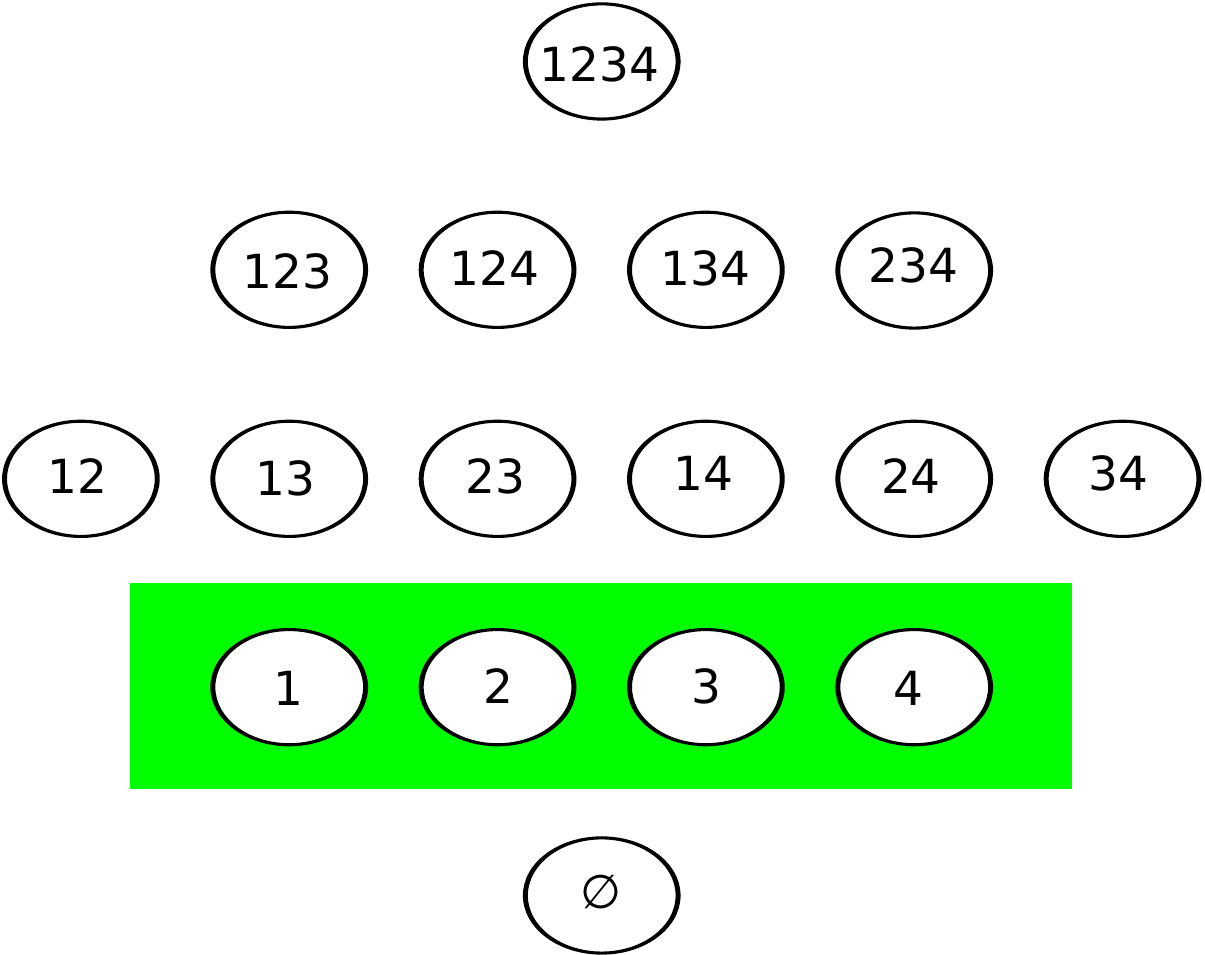} \hspace{-0.07in} &
\hspace{-0.06in} \includegraphics[width=0.23\textwidth]{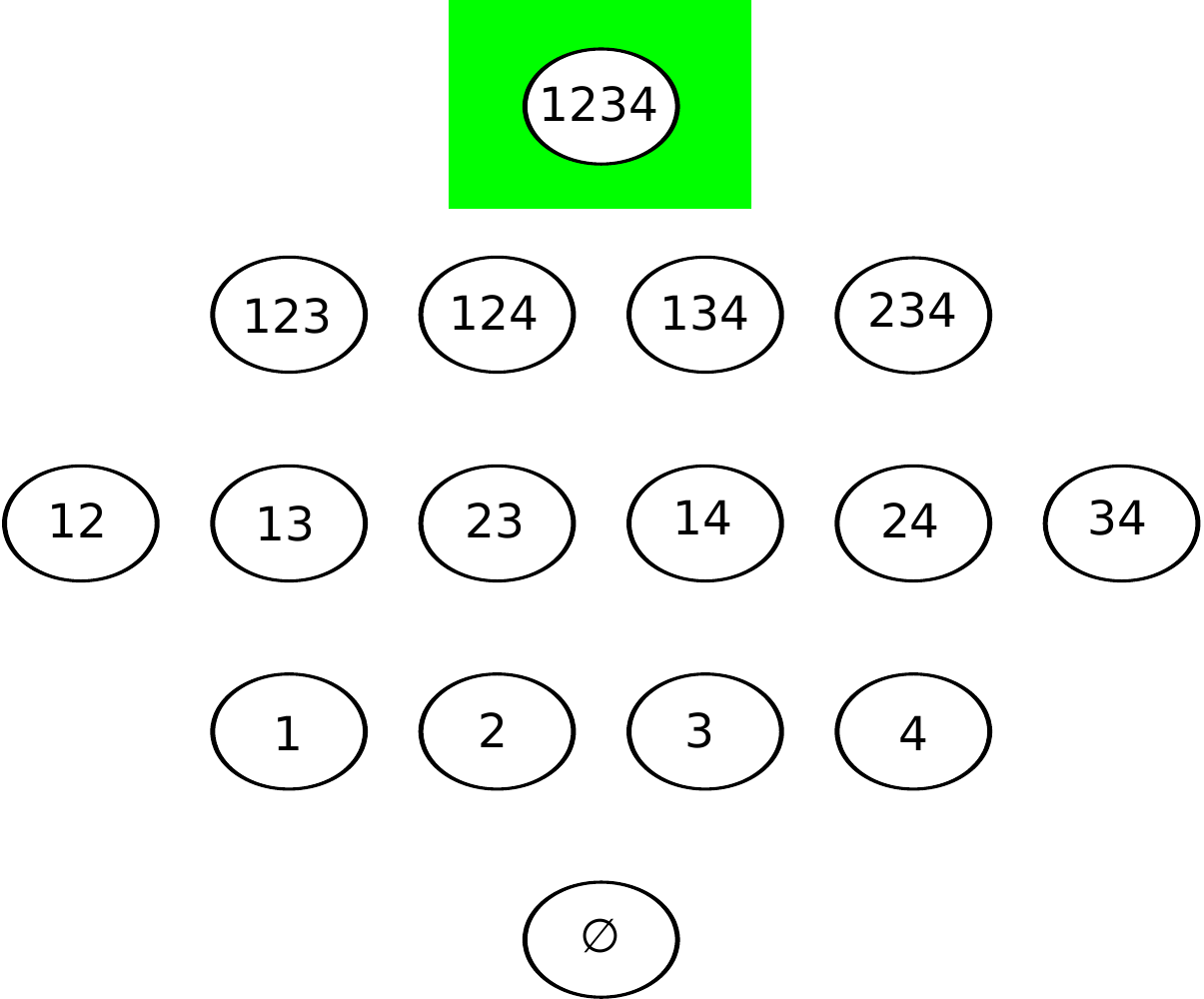} \hspace{-0.07in} &
\hspace{-0.06in} \includegraphics[height=0.19\textwidth]{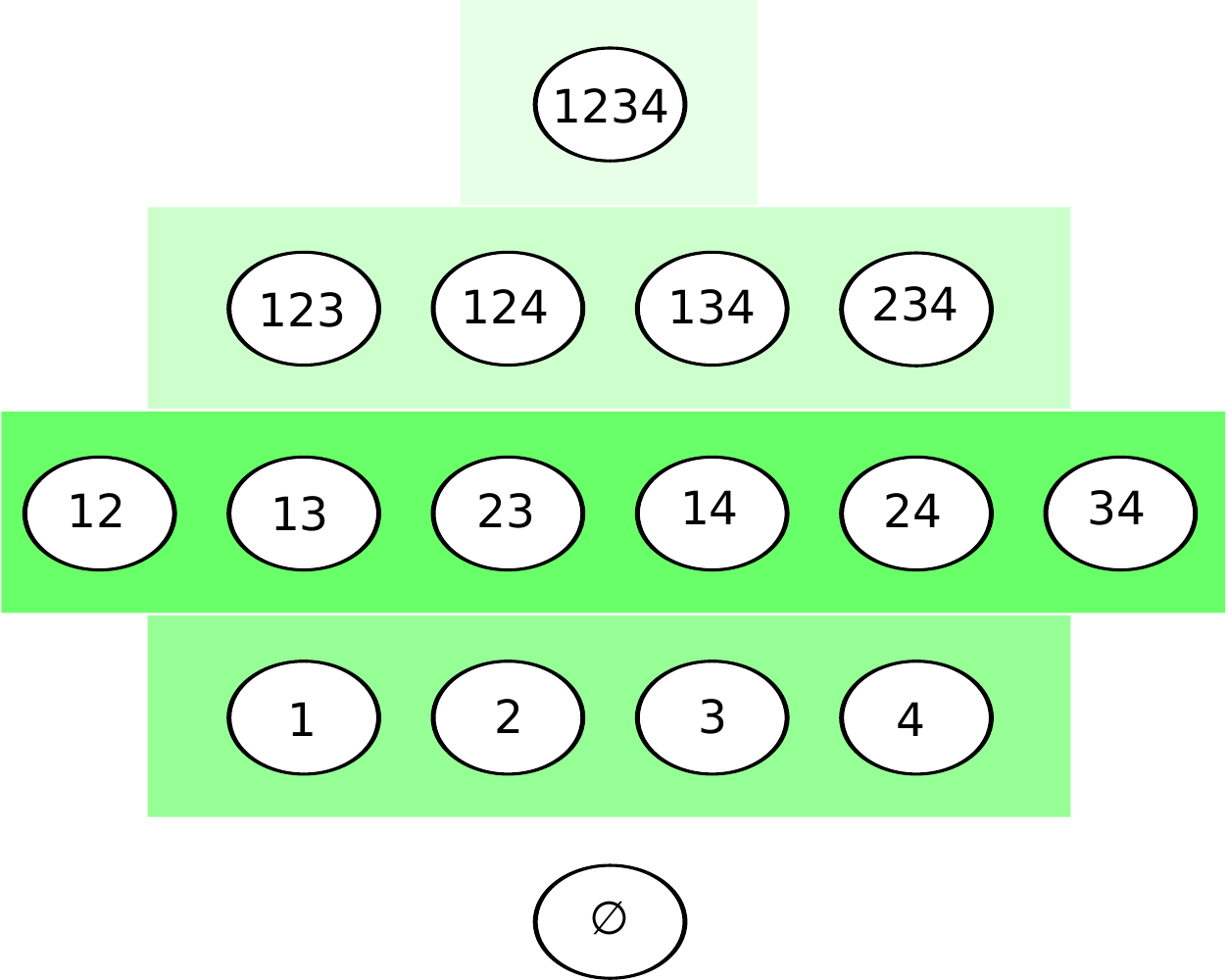} \\
HKL kernel & GP-GAM kernel & Squared-exp GP & Additive GP kernel\\
 & & kernel & \\
\end{tabular}
\caption{
A comparison of different models.  Nodes represent different interaction terms, ranging from first-order to fourth-order interactions.  Far left:  HKL can select a hull of interaction terms, but must use a pre-determined weighting over those terms.  Far right: the additive GP model can weight each order of interaction seperately.  Neither the HKL nor the additive model dominate one another in terms of flexibility, however the GP-GAM and the SE-GP are special cases of additive GPs. }
\label{hulls-figure}
\end{figure}

Bach\cite{DBLP:journals/corr/abs-0909-0844} uses a regularized optimization framework to learn a weighted sum over an exponential number of kernels which can be computed in polynomial time.  The subsets of kernels considered by this method are restricted to be a \textit{hull} of kernels.\footnote{In the setting we are considering in this paper, a hull can be defined as a subset of all terms such that if term $\prod_{j \in J} k_j(\bf x, x')$ is included in the subset, then so are all terms $\prod_{j \in J / i} k_j(\bf x, x')$, for all $i \in J$.  For details, see \cite{DBLP:journals/corr/abs-0909-0844}.}
Given each dimension's kernel, and a pre-defined weighting over all terms, HKL performs model selection by searching over hulls of interaction terms.
In \cite{DBLP:journals/corr/abs-0909-0844}, Bach also fixes the relative weighting between orders of interaction with a single term $\alpha$, computing the sum over all orders by:
\begin{equation}
\label{eqn:uniform}
k_{a}({\bf x, x'}) = v_D^2 \prod_{d=1}^D \left(1 + \alpha k_{d}(x_{d}, x_{d}') \right)
\end{equation}
which has computational complexity $O(D)$.  However, this formulation forces the weight of all $n$th order terms to be weighted by $\alpha^n$.

Figure \ref{hulls-figure} contrasts the HKL hull-selection method with the Additive GP hyperparameter-learning method. Neither method dominates the other in flexibility.  The main difficulty with the approach of \cite{DBLP:journals/corr/abs-0909-0844} is that hyperparameters are hard to set other than by cross-validation.  In contrast, our method optimizes the hyperparameters of each dimension's base kernel, as well as the relative weighting of each order of interaction.

\subsection{ANOVA Procedures}

Vapnik \cite{vapnik1998statistical} introduces the support vector ANOVA decomposition, which has the same form as our additive kernel.  However, they recommend approximating the sum over all $D$ orders with only one term ``of appropriate order'', presumably because of the difficulty of setting the hyperparameters of an SVM. Stitson et al.\cite{stitson1999support} performed experiments which favourably compared the support vector ANOVA decomposition to polynomial and spline kernels.  They too allowed only one order to be active, and set hyperparameters by cross-validation.
%

A closely related procedure from the statistics literature is smoothing-splines ANOVA (SS-ANOVA)\cite{wahba1990spline}. An SS-ANOVA model is estimated as a weighted sum of splines along each dimension, plus a sum of splines over all pairs of dimensions, all triplets, etc, with each individual interaction term having a separate weighting parameter.  Because the number of terms to consider grows exponentially in the order, in practice, only terms of first and second order are usually considered.  Learning in SS-ANOVA is usually done via penalized-maximum likelihood with a fixed sparsity hyperparameter.

In contrast to these procedures, our method can easily include all $D$ orders of interaction, each weighted by a separate hyperparameter. As well, we can learn kernel hyperparameters individually per input dimension, allowing automatic relevance determination to operate.

\subsection{Non-local Interactions}

%
By far the most popular kernels for regression and classification tasks on continuous data are the squared exponential (Gaussian) kernel, and the Mat\'{e}rn kernels.  These kernels depend only on the scaled Euclidean distance between two points, both having the form: $ k({\bf x, x'}) = f(\sum_{d=1}^D \left( x_{d} - x_{d}' \right)^2 / l_d^2)$.
Bengio et al.\cite{bengio2006curse} argue that models based on squared-exponential kernels are particularily susceptible to the \textit{curse of dimensionality}.  They emphasize that the locality of the kernels means that these models cannot capture non-local structure.  They argue that many functions that we care about have such structure.  Methods based solely on local kernels will require training examples at all combinations of relevant inputs.

\begin{figure}[h]
\centering
\begin{tabular}{cccc}
\hspace{-0.25in} \includegraphics[width=0.27\textwidth]{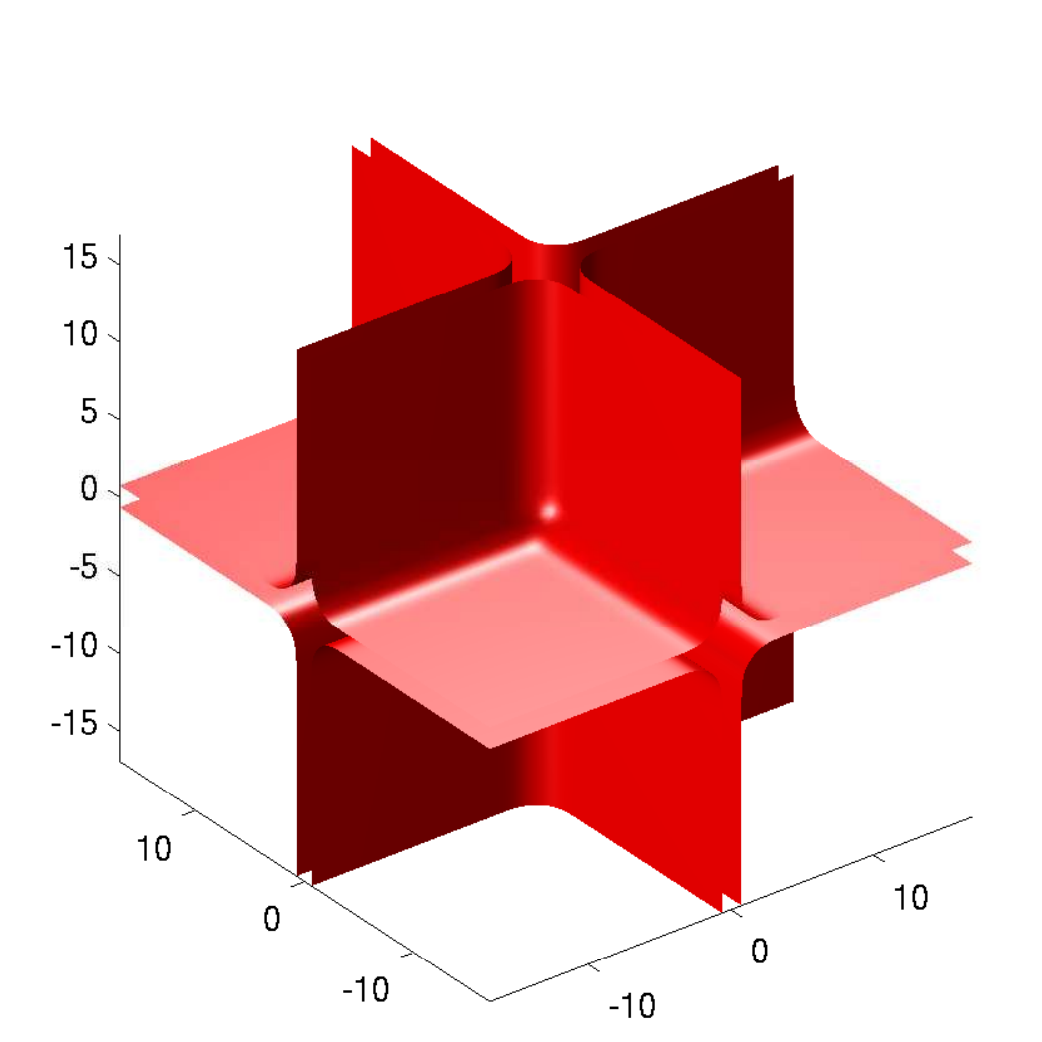} &
\hspace{-0.25in} \includegraphics[width=0.27\textwidth]{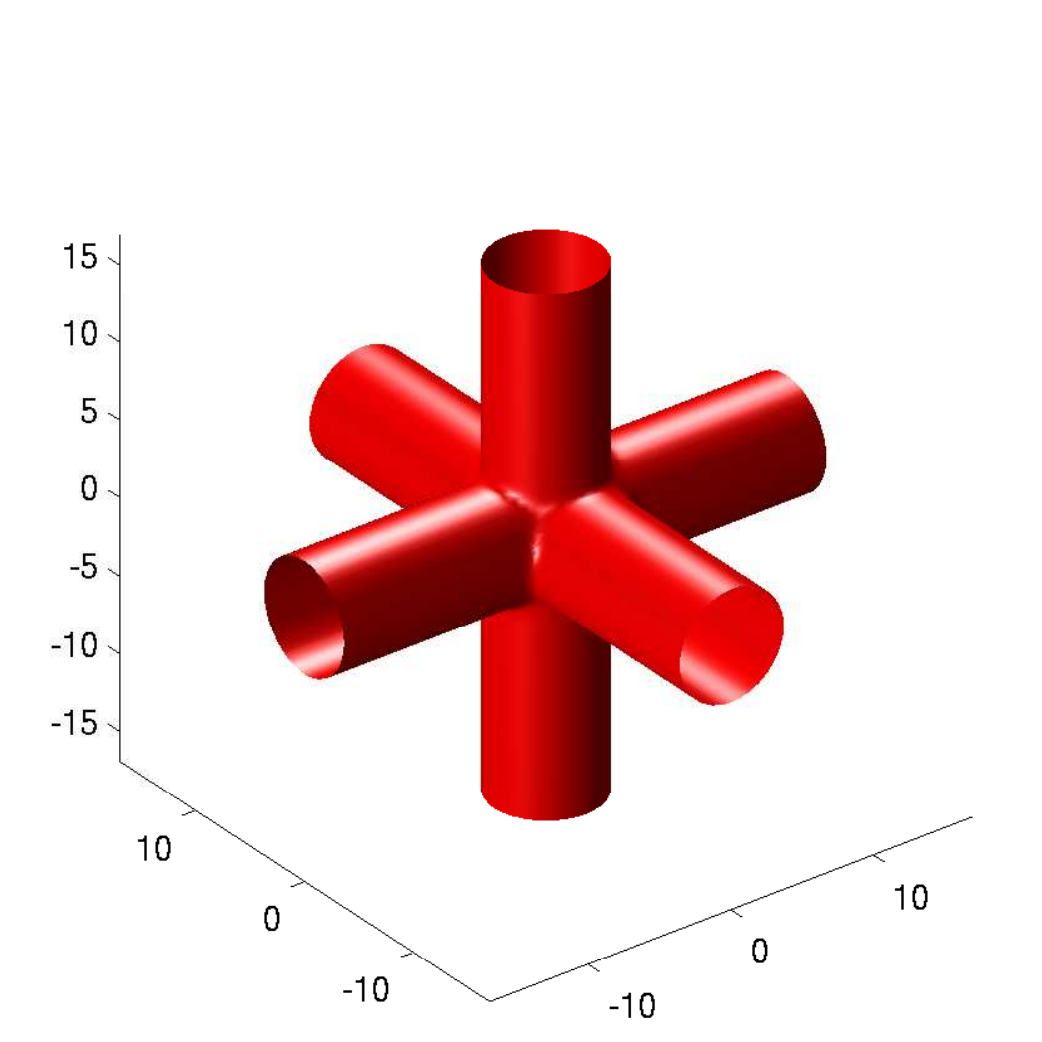} &
\hspace{-0.25in} \includegraphics[width=0.27\textwidth]{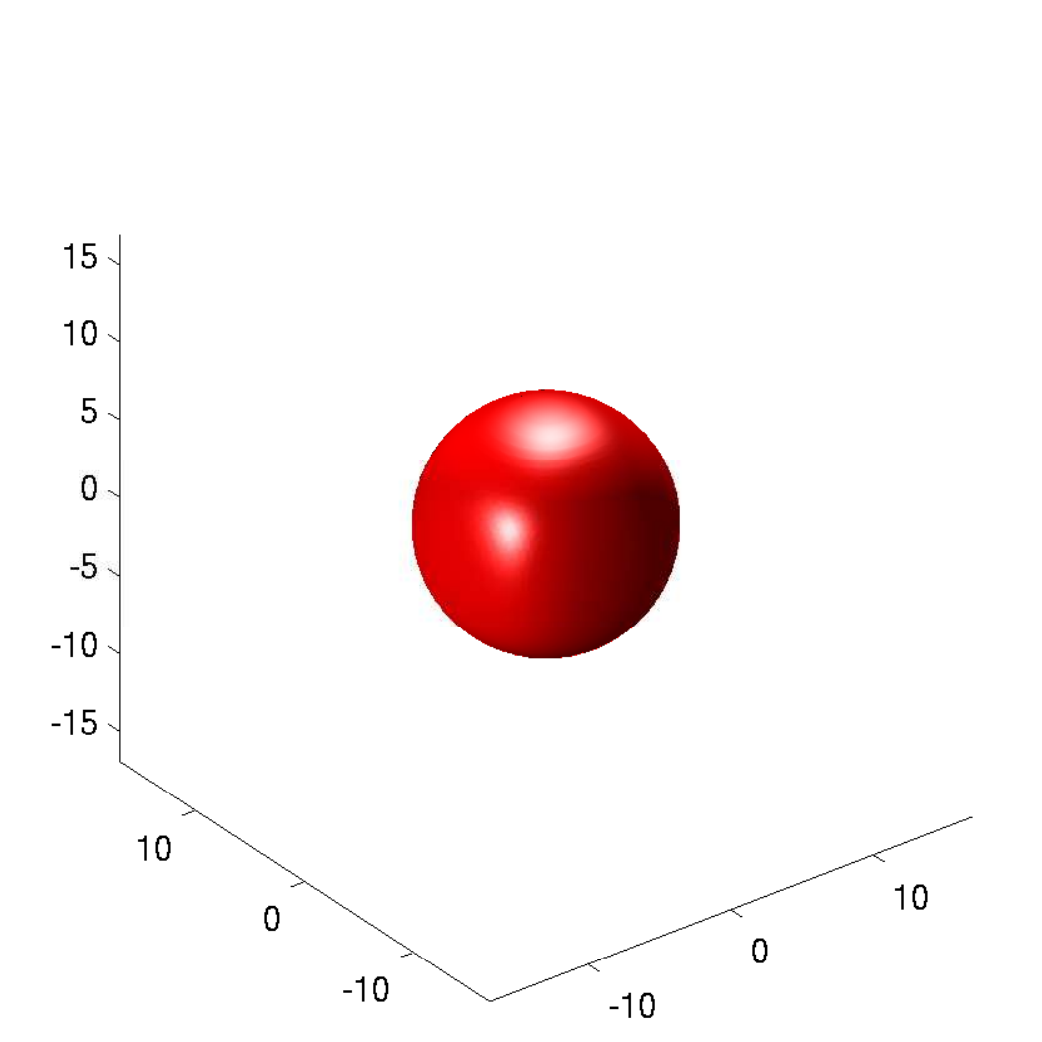} & 
\hspace{-0.25in} \includegraphics[width=0.27\textwidth]{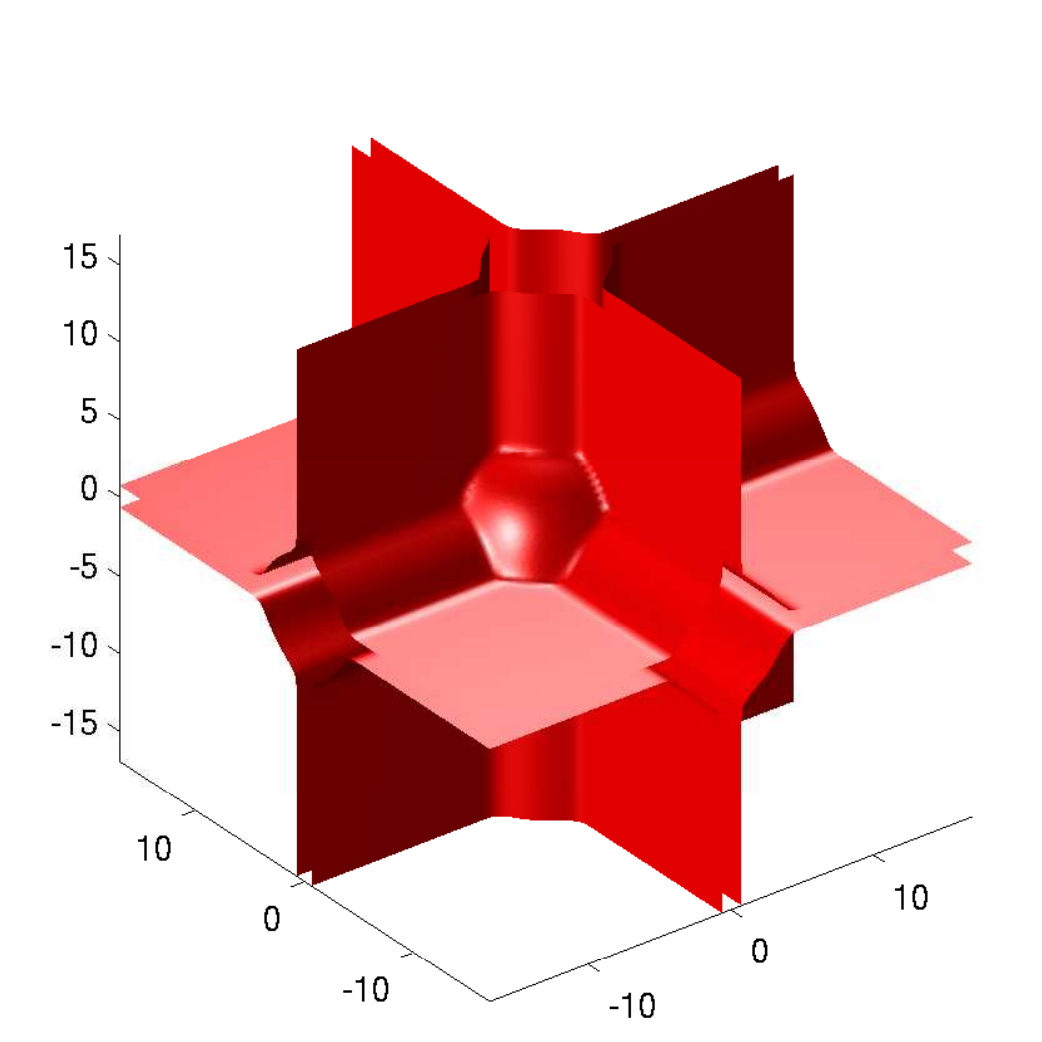}\\
1st order interactions & 2nd order interactions & 3rd order interactions & All interactions \\
$k_1 + k_2 + k_3$ & $k_1k_2 + k_2k_3 + k_1k_3$ & $k_1k_2k_3$ & \\
& & (Squared-exp kernel) & (Additive kernel)\\
\end{tabular}
\caption{Isocontours of additive kernels in 3 dimensions.  The third-order kernel only considers nearby points relevant, while the lower-order kernels allow the output to depend on distant points, as long as they share one or more input value.}
\label{fig:kernels3d}
\end{figure}

Additive kernels have a much more complex structure, and allow extrapolation based on distant parts of the input space, without spreading the mass of the kernel over the whole space.  For example, additive kernels of the second order allow strong non-local interactions between any points which are similar in any two input dimensions.
Figure \ref{fig:kernels3d} provides a geometric comparison between squared-exponential kernels and additive kernels in 3 dimensions.

\section{Experiments}

\subsection{Synthetic Data}

Because additive kernels can discover non-local structure in data, they are exceptionally well-suited to problems where local interpolation fails.  
\begin{figure}[h]
\centering
\begin{tabular}{cccc}
\hspace{-0.1in}\includegraphics[width=0.24\textwidth]{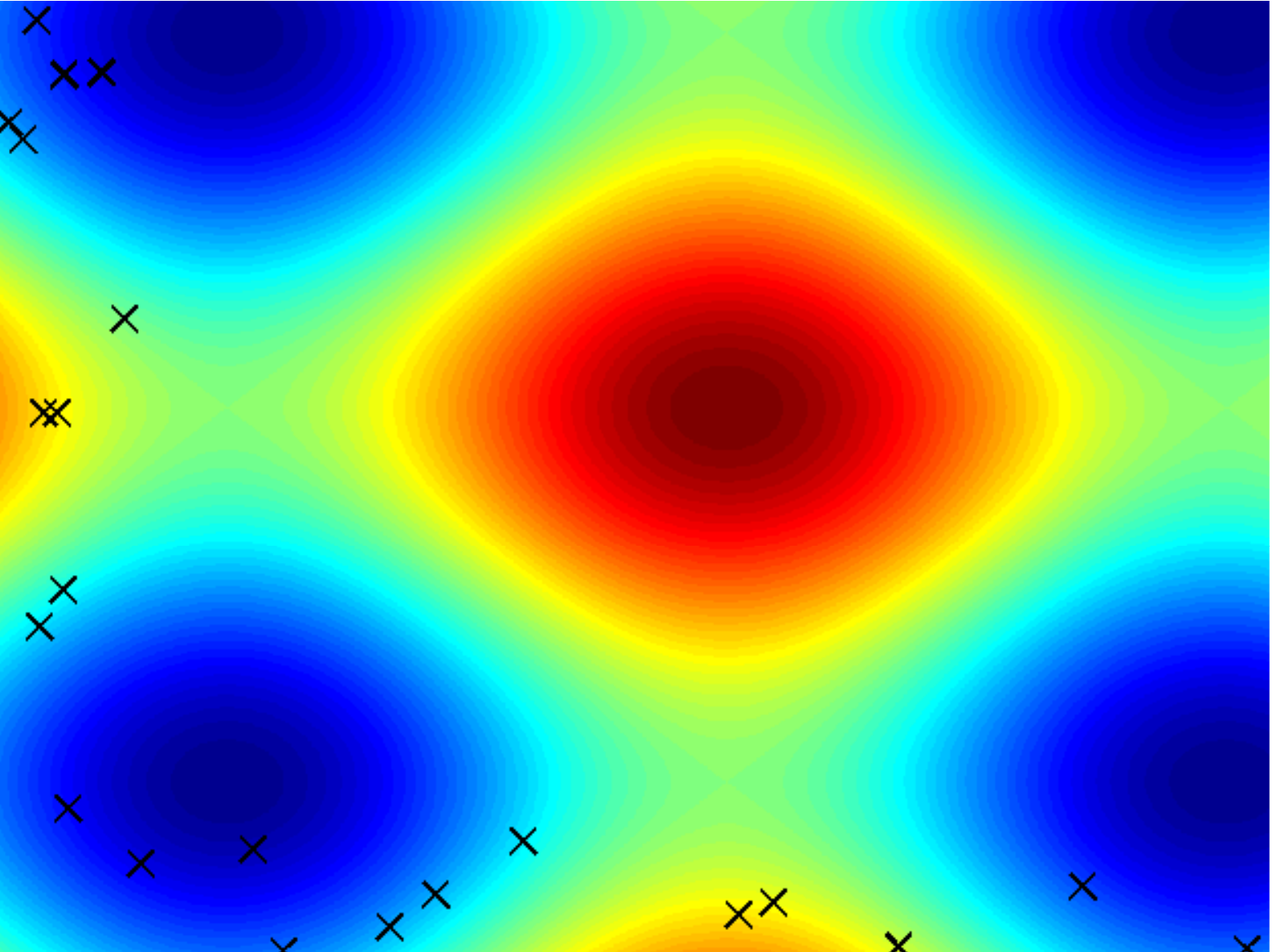} &
\hspace{-0.1in}\includegraphics[width=0.24\textwidth]{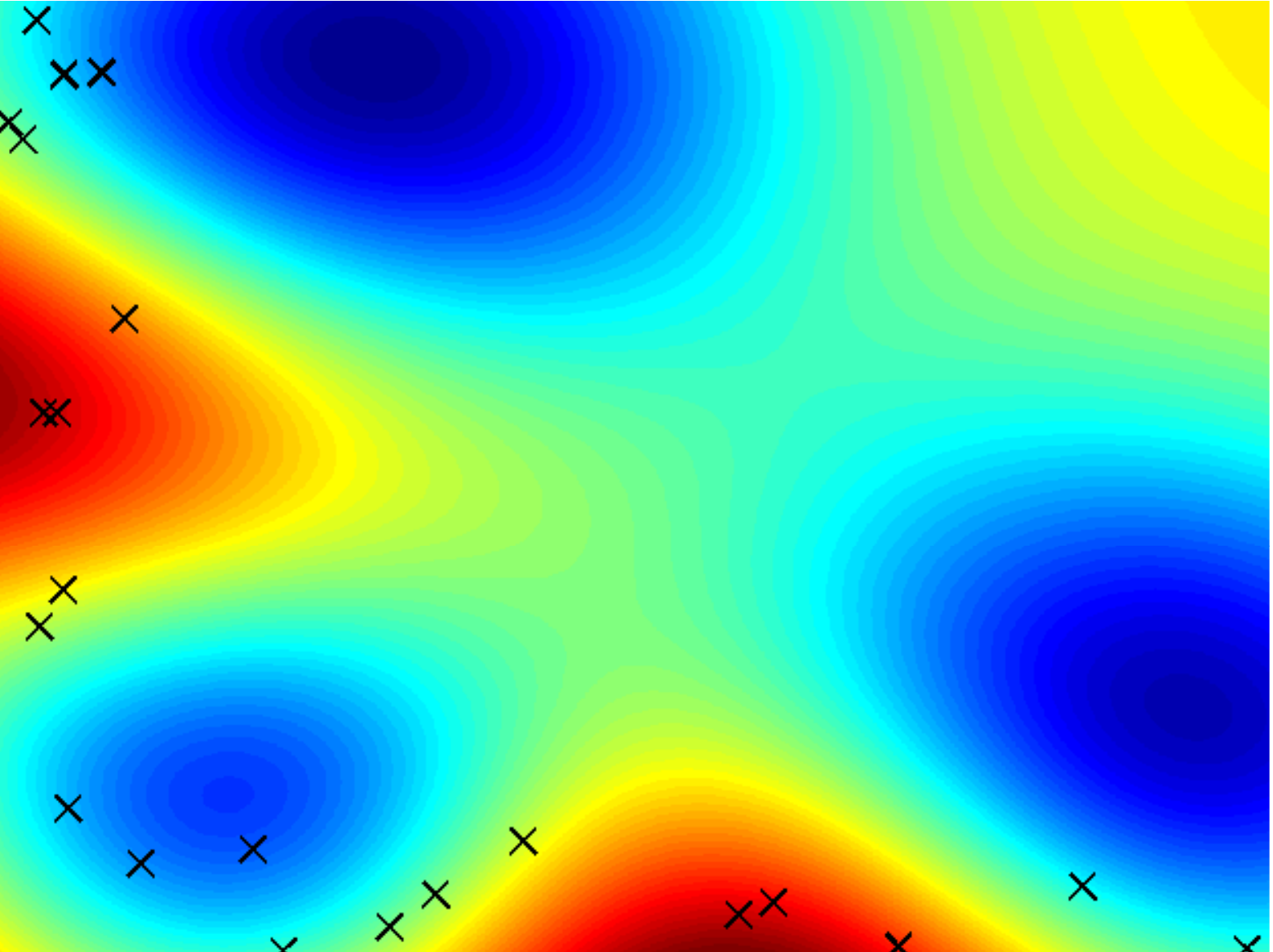}&
\hspace{-0.1in}\includegraphics[width=0.24\textwidth]{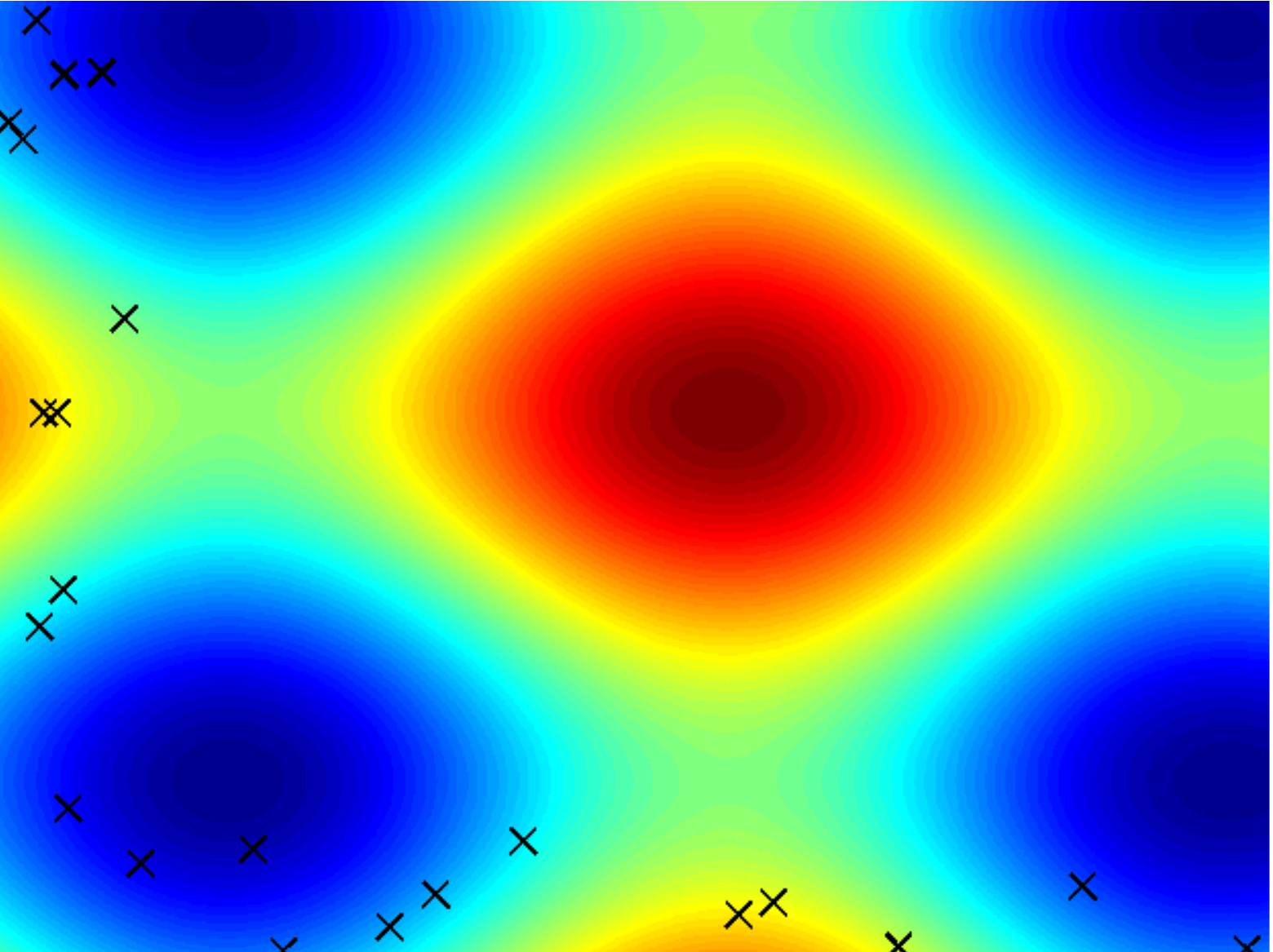}& 
\hspace{-0.1in}\includegraphics[width=0.24\textwidth]{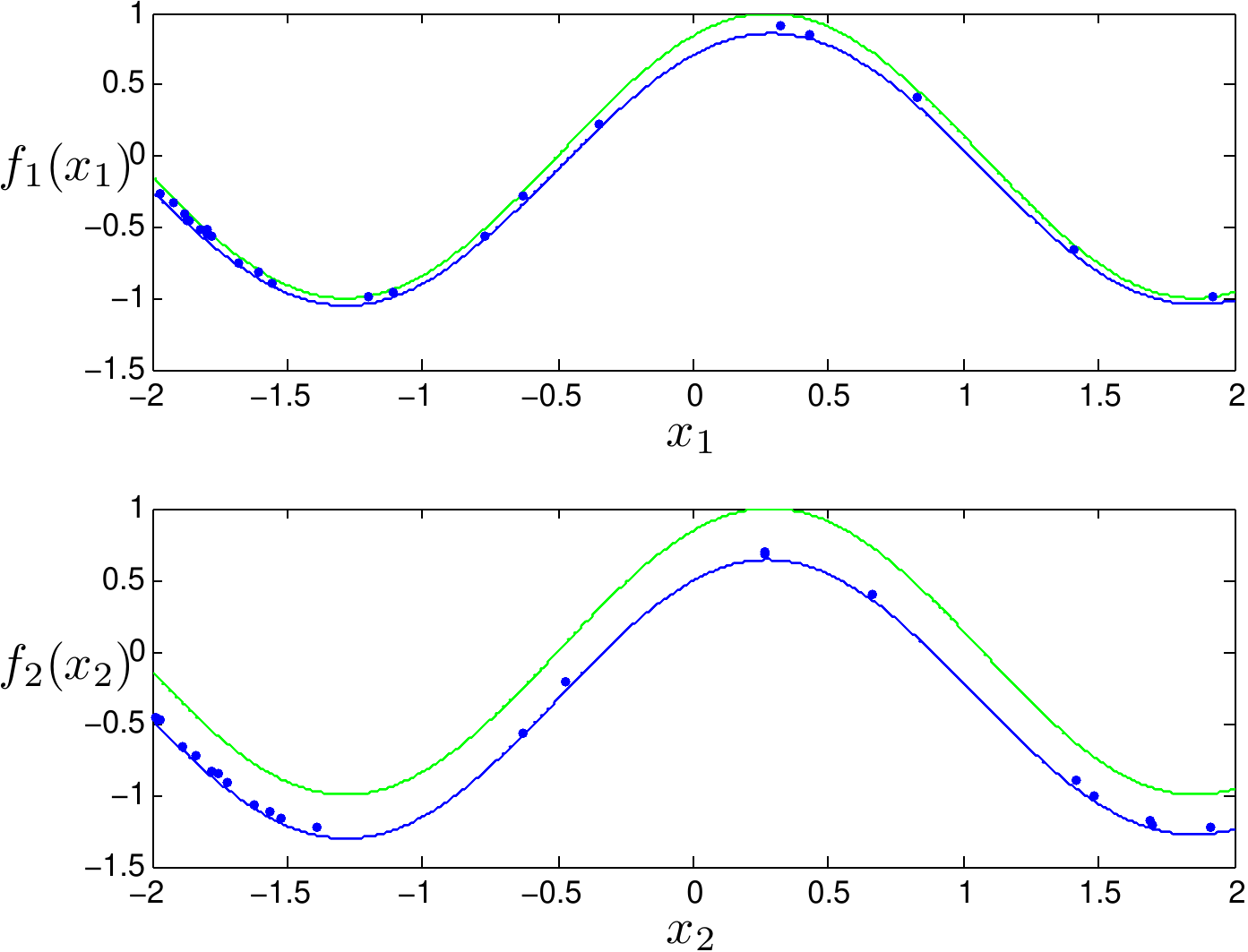}\\ 
True Function & Squared-exp GP & Additive GP & Additive GP \\
 \& data locations & posterior mean & posterior mean & 1st-order functions\\
\end{tabular}
\caption{Long-range inference in functions with additive structure.
}
\label{fig:synth2d}
\end{figure}
Figure \ref{fig:synth2d} shows a dataset which demonstrates this feature of additive GPs, consisting of data drawn from a sum of two axis-aligned sine functions.  The training set is restricted to a small, L-shaped area; the test set contains a peak far from the training set locations.  The additive GP recovered both of the original sine functions (shown in green), and inferred correctly that most of the variance in the function comes from first-order interactions.  The ability of additive GPs to discover long-range structure suggests that this model may be well-suited to deal with covariate-shift problems.

\subsection{Experimental Setup}

On a diverse collection of datasets, we compared five different models.  In the results tables below, GP Additive refers to a GP using the additive kernel with squared-exp base kernels.  For speed, we limited the maximum order of interaction in the additive kernels to 10.  GP-GAM denotes an additive GP model with only first-order interactions.  GP Squared-Exp is a GP model with a squared-exponential ARD kernel.  HKL\footnote{Code for HKL available at \texttt{http://www.di.ens.fr/\textasciitilde fbach/hkl/}} was run using the all-subsets kernel, which corresponds to the same set of kernels as considered by the additive GP with a squared-exp base kernel.     

For all GP models, we fit hyperparameters by the standard method of maximizing training-set marginal likelihood, using L-BFGS \cite{nocedal1980updating} for 500 iterations, allowing five random restarts.  In addition to learning kernel hyperparameters, we fit a constant mean function to the data.
In the classification experiments, GP inference was done using Expectation Propagation \cite{minka2001expectation}.
 
\subsection{Results}

Tables \ref{tbl:Regression Mean Squared Error}, \ref{tbl:Regression Negative Log Likelihood}, \ref{tbl:Classification Percent Error} and \ref{tbl:Classification Negative Log Likelihood} show mean performance across 10 train-test splits.  Because HKL does not specify a noise model, it could not be included in the likelihood comparisons.

\begin{table}[h!]
\caption{{\small
Regression Mean Squared Error
}}
\label{tbl:Regression Mean Squared Error}
\begin{center}
\begin{tabular}{l | r r r r r}
Method & \rotatebox{0}{ bach  }  & \rotatebox{0}{ concrete  }  & \rotatebox{0}{ pumadyn-8nh }  & \rotatebox{0}{ servo }  & \rotatebox{0}{ housing }  \\ \hline
Linear Regression & $1.031$ & $0.404$ & $0.641$ & $0.523$ & $0.289$ \\
GP GAM & $1.259$ & $0.149$ & $0.598$ & $0.281$ & $0.161$ \\
HKL & $\mathbf{0.199}$ & $0.147$ & $0.346$ & $0.199$ & $0.151$ \\
GP Squared-exp & $\mathbf{0.045}$ & $0.157$ & $\mathbf{0.317}$ & $\mathbf{0.126}$ & $\mathbf{0.092}$ \\
GP Additive & $\mathbf{0.045}$ & $\mathbf{0.089}$ & $\mathbf{0.316}$ & $\mathbf{0.110}$ & $\mathbf{0.102}$ \\
\end{tabular}
\end{center}
\end{table}
\begin{table}[h!]
\caption{{\small
Regression Negative Log Likelihood
}}
\label{tbl:Regression Negative Log Likelihood}
\begin{center}
\begin{tabular}{l | r r r r r}
Method & \rotatebox{0}{ bach  }  & \rotatebox{0}{ concrete  }  & \rotatebox{0}{ pumadyn-8nh }  & \rotatebox{0}{ servo }  & \rotatebox{0}{ housing }  \\ \hline
Linear Regression & $2.430$ & $1.403$ & $1.881$ & $1.678$ & $1.052$ \\
GP GAM & $1.708$ & $0.467$ & $1.195$ & $0.800$ & $0.457$ \\
GP Squared-exp & $\mathbf{-0.131}$ & $0.398$ & $\mathbf{0.843}$ & $0.429$ & $\mathbf{0.207}$ \\
GP Additive & $\mathbf{-0.131}$ & $\mathbf{0.114}$ & $\mathbf{0.841}$ & $\mathbf{0.309}$ & $\mathbf{0.194}$ \\
\end{tabular}
\end{center}
\end{table}
\begin{table}[h!]
\caption{{\small
Classification Percent Error
}}
\label{tbl:Classification Percent Error}
\begin{center}
\begin{tabular}{l | r r r r r r}
Method & \rotatebox{0}{ breast }  & \rotatebox{0}{ pima }  & \rotatebox{0}{ sonar }  & \rotatebox{0}{ ionosphere }  & \rotatebox{0}{ liver }  & \rotatebox{0}{ heart }  \\ \hline
Logistic Regression & $7.611$ & $24.392$ & $26.786$ & $16.810$ & $45.060$ & $\mathbf{16.082}$ \\
GP GAM & $\mathbf{5.189}$ & $\mathbf{22.419}$ & $\mathbf{15.786}$ & $\mathbf{8.524}$ & $\mathbf{29.842}$ & $\mathbf{16.839}$ \\
HKL & $\mathbf{5.377}$ & $24.261$ & $\mathbf{21.000}$ & $9.119$ & $\mathbf{27.270}$ & $\mathbf{18.975}$ \\
GP Squared-exp & $\mathbf{4.734}$ & $\mathbf{23.722}$ & $\mathbf{16.357}$ & $\mathbf{6.833}$ & $\mathbf{31.237}$ & $\mathbf{20.642}$ \\
GP Additive & $\mathbf{5.566}$ & $\mathbf{23.076}$ & $\mathbf{15.714}$ & $\mathbf{7.976}$ & $\mathbf{30.060}$ & $\mathbf{18.496}$ \\
\end{tabular}
\end{center}
\end{table}
\begin{table}[h!]
\caption{{\small
Classification Negative Log Likelihood
}}
\label{tbl:Classification Negative Log Likelihood}
\begin{center}
\begin{tabular}{l | r r r r r r}
Method & \rotatebox{0}{ breast }  & \rotatebox{0}{ pima }  & \rotatebox{0}{ sonar }  & \rotatebox{0}{ ionosphere }  & \rotatebox{0}{ liver }  & \rotatebox{0}{ heart }  \\ \hline
Logistic Regression & $0.247$ & $0.560$ & $4.609$ & $0.878$ & $0.864$ & $0.575$ \\
GP GAM & $\mathbf{0.163}$ & $\mathbf{0.461}$ & $\mathbf{0.377}$ & $\mathbf{0.312}$ & $\mathbf{0.569}$ & $\mathbf{0.393}$ \\
GP Squared-exp & $\mathbf{0.146}$ & $0.478$ & $\mathbf{0.425}$ & $\mathbf{0.236}$ & $\mathbf{0.601}$ & $0.480$ \\
GP Additive & $\mathbf{0.150}$ & $\mathbf{0.466}$ & $\mathbf{0.409}$ & $\mathbf{0.295}$ & $\mathbf{0.588}$ & $\mathbf{0.415}$ \\
\end{tabular}
\end{center}
\end{table}

The model with best performance on each dataset is in bold, along with all other models that were not significantly different under a paired t-test. The additive model never performs significantly worse than any other model, and sometimes performs significantly better than all other models.  The difference between all methods is larger in the case of regression experiments. The performance of HKL is consistent with the results in \cite{DBLP:journals/corr/abs-0909-0844}, performing competitively but slightly worse than SE-GP.

The additive GP performed best on datasets well-explained by low orders of interaction, and approximately as well as the SE-GP model on datasets which were well explained by high orders of interaction (see table \ref{tbl:all_orders}).
Because the additive GP is a superset of both the GP-GAM model and the SE-GP model, instances where the additive GP performs slightly worse are presumably due to over-fitting, or due to the hyperparameter optimization becoming stuck in a local maximum. 
Additive GP performance can be expected to benefit from integrating out the kernel hyperparameters.

\section{Conclusion}

We present additive Gaussian processes: a simple family of models which generalizes two widely-used classes of models.  Additive GPs also introduce a tractable new type of structure into the GP framework.   Our experiments indicate that such additive structure is present in real datasets, allowing our model to perform better than standard GP models.  In the case where no such structure exists, our model can recover arbitrarily flexible models, as well.

In addition to improving modeling efficacy, the additive GP also improves model interpretability:  the order variance hyperparameters indicate which sorts of structure are present in our model.

Compared to HKL, which is the only other tractable procedure able to capture the same types of structure, our method benefits from being able to learn individual kernel hyperparameters, as well as the weightings of different orders of interaction.  Our experiments show that additive GPs are a state-of-the-art regression model.


\subsubsection*{Acknowledgments}
The authors would like to thank John J. Chew and Guillaume Obozonksi for their helpful comments.
\bibliographystyle{unsrt}
\bibliography{additive}

\begin{thebibliography}{10}

\bibitem{nelder1972generalized}
J.A. Nelder and R.W.M. Wedderburn.
\newblock Generalized linear models.
\newblock {\em Journal of the Royal Statistical Society. Series A (General)},
  135(3):370--384, 1972.

\bibitem{hastie1990generalized}
T.J. Hastie and R.J. Tibshirani.
\newblock {\em Generalized additive models}.
\newblock Chapman \& Hall/CRC, 1990.

\bibitem{wahba1990spline}
G.~Wahba.
\newblock {\em {Spline models for observational data}}.
\newblock Society for Industrial Mathematics, 1990.

\bibitem{DBLP:journals/corr/abs-0909-0844}
Francis Bach.
\newblock High-dimensional non-linear variable selection through hierarchical
  kernel learning.
\newblock {\em CoRR}, abs/0909.0844, 2009.

\bibitem{rasmussen38gaussian}
C.E. Rasmussen and CKI Williams.
\newblock {Gaussian Processes for Machine Learning}.
\newblock {\em The MIT Press, Cambridge, MA, USA}, 2006.

\bibitem{plate1999accuracy}
T.A. Plate.
\newblock {Accuracy versus interpretability in flexible modeling: Implementing
  a tradeoff using Gaussian process models}.
\newblock {\em Behaviormetrika}, 26:29--50, 1999.

\bibitem{macdonald1998symmetric}
I.G. Macdonald.
\newblock {\em {Symmetric functions and Hall polynomials}}.
\newblock Oxford University Press, USA, 1998.

\bibitem{stanley2001enumerative}
R.P. Stanley.
\newblock {\em {Enumerative combinatorics}}.
\newblock Cambridge University Press, 2001.

\bibitem{kaufman2010bayesian}
C.G. Kaufman and S.R. Sain.
\newblock Bayesian functional anova modeling using {G}aussian process prior
  distributions.
\newblock {\em Bayesian Analysis}, 5(1):123--150, 2010.

\bibitem{christoudias2009bayesian}
M.~Christoudias, R.~Urtasun, and T.~Darrell.
\newblock {Bayesian localized multiple kernel learning}.
\newblock {\em Technical report}, 2009.

\bibitem{vapnik1998statistical}
V.N. Vapnik.
\newblock {\em {Statistical learning theory}}, volume~2.
\newblock Wiley New York, 1998.

\bibitem{stitson1999support}
M.~Stitson, A.~Gammerman, V.~Vapnik, V.~Vovk, C.~Watkins, and J.~Weston.
\newblock {Support vector regression with ANOVA decomposition kernels}.
\newblock {\em Advances in kernel methods: Support vector learning}, pages
  285--292, 1999.

\bibitem{bengio2006curse}
Y.~Bengio, O.~Delalleau, and N.~Le~Roux.
\newblock {The curse of highly variable functions for local kernel machines}.
\newblock {\em Advances in neural information processing systems}, 18, 2006.

\bibitem{nocedal1980updating}
J.~Nocedal.
\newblock Updating quasi-newton matrices with limited storage.
\newblock {\em Mathematics of computation}, 35(151):773--782, 1980.

\bibitem{minka2001expectation}
T.P. Minka.
\newblock Expectation propagation for approximate {B}ayesian inference.
\newblock In {\em Uncertainty in Artificial Intelligence}, volume~17, pages
  362--369, 2001.

\end{thebibliography}

\end{document}